\pdfoutput=1

\documentclass[11pt]{article}

\usepackage[final]{acl}

\usepackage{times}
\usepackage{latexsym}

\usepackage[T1]{fontenc}

\usepackage[utf8]{inputenc}

\usepackage{microtype}

\usepackage{inconsolata}

\usepackage{graphicx}

\usepackage{amsmath,amsfonts,bm}

\def\eqref#1{equation~\ref{#1}}

\def\1{\bm{1}}

\DeclareMathAlphabet{\mathsfit}{\encodingdefault}{\sfdefault}{m}{sl}
\SetMathAlphabet{\mathsfit}{bold}{\encodingdefault}{\sfdefault}{bx}{n}

\DeclareMathOperator*{\argmin}{arg\,min}

\usepackage{soul}
\usepackage{url}
\usepackage[utf8]{inputenc} 
\usepackage{caption}
\usepackage{graphicx}
\usepackage{amsmath}
\usepackage{hyperref}
\usepackage[capitalize,noabbrev]{cleveref}
\makeatletter
\AddToHook{cmd/appendix/before}{\def\cref@section@alias{appendix}}
\AddToHook{cmd/appendix/before}{\def\cref@subsection@alias{appendix}}
\makeatother

\usepackage{amsthm}
\usepackage{booktabs}
\usepackage{algorithm}
\usepackage{algorithmicx}
\usepackage{algpseudocode}
\usepackage[switch]{lineno}

\usepackage{amssymb}
\usepackage{array}
\usepackage{tabularx}
\usepackage{setspace}
\usepackage{enumitem}

\usepackage{graphicx}
\usepackage{float} 
\usepackage{subfigure}
\usepackage{adjustbox}

\usepackage{float}
\usepackage{multirow}
\usepackage{wrapfig}
\usepackage{colortbl}
\usepackage{makecell}
\usepackage{bbding}

\title{Optimizing Cross-Client Domain Coverage for Federated Instruction Tuning of Large Language Models}

\author{
 \textbf{Zezhou Wang\textsuperscript{1}},
 \textbf{Yaxin Du\textsuperscript{2}},
 \textbf{Xingjun Ma\textsuperscript{3}}, \\
 \textbf{Yugang Jiang\textsuperscript{3}},
 \textbf{Zhuzhong Qian\textsuperscript{1 *}},
 \textbf{Siheng Chen\textsuperscript{2 *}},
\\
\textsuperscript{1}Nanjing University, 
\textsuperscript{2}Shanghai Jiao Tong University, \\
\textsuperscript{3}Fudan University \\
\small{\texttt{\{zzw.cs@smail, qzz@\}nju.edu.cn}, \texttt{\{dorothydu, sihengc\}@sjtu.edu.cn}, \texttt{\{xingjunma, ygj\}@fudan.edu.cn}}
}

\begin{document}
\maketitle
\begingroup
\renewcommand\thefootnote{$*$}
\footnotetext{Corresponding author.}
\endgroup
\begin{abstract}

Federated domain-specific instruction tuning (FedDIT) for large language models (LLMs) aims to enhance performance in specialized domains using distributed private and limited data, yet identifying key performance drivers and optimal augmentation strategies remains challenging. We empirically establish that cross-client domain coverage, rather than data heterogeneity, is the pivotal factor. We then introduce FedDCA, an algorithm that explicitly maximizes this coverage through diversity-oriented client center selection and retrieval-based augmentation, constructing diverse, non-redundant cross-client instruction sets. Extensive experiments across multiple domains demonstrate FedDCA's superiority over eleven baselines, achieving performance gains of up to 29.19\% and domain coverage improvements of 4.82\%-21.36\%. FedDCA maintains its effectiveness in diverse and challenging scenarios, including data selection, held-out settings where task-specific public data is scarce and various data heterogeneity, with manageable privacy risks. This work clarifies critical FedDIT dynamics and presents FedDCA as an effective, privacy-preserving, and scalable solution for advancing domain-specific LLM tuning.
\end{abstract}

\section{Introduction}

Recently, federated instruction tuning (FedIT) has gained attention as a novel approach that leverages the principles of federated learning (FL) to facilitate collaborative training of large language models (LLM) in distributed environments while maintaining the confidentiality of private data \cite{2017-McMahan-Communicationefficient-AIS,2024-Ye-OpenFedLLM-,2024-Zhang-Building-}. This methodology allows for the exchange of model parameters among distributed data holders, thereby achieving a careful balance between privacy preservation and efficient model optimization. Despite the establishment of various FedIT frameworks \cite{2024-Ye-OpenFedLLM-,2023-Kuang-FederatedScopeLLM-,2024-Zhang-Building-}, existing literature has not adequately addressed the practical challenges that Federated Domain-specific Instruction Tuning (FedDIT) may encounter in real-world applications. For instance, FedIT generally necessitates a sufficient amount of instruction data for fine-tuning, which is often a shortage in domain-specific fine-tuning contexts \cite{2024-Zhang-FewFedPIT-}.

We explore Federated Domain-specific Instruction Tuning (FedDIT), an innovative approach that harnesses federated learning (FL) to unlock the potential of Large Language Models (LLMs) in specialized domains. A significant hurdle in deploying such models arises when multiple entities hold valuable but limited and privacy-sensitive data—a common scenario in fields like medical diagnosis \cite{Guan2023FederatedLFMedical,OCIF-Hu2023} and financial risk assessment \cite{Abadi2024StarlitPF}. These entities require collaborative training of domain-specific LLMs without direct data sharing \cite{Xu2024DoFITDF,2024-Zhang-FewFedPIT-}. Our findings underscore a key limitation: exclusive reliance on local, in-domain data, despite its quality, often results in subpar performance due to insufficient scale (see \cref{tab:performance}). FedDIT tackles this by integrating carefully designed instruction augmentation strategies (detailed in \cref{sec:related_work}) that expand local datasets while upholding privacy. This augmentation is not merely a supplement but a crucial enabler for robust instruction tuning and for averting performance decline. Focusing on practical and high-quality augmentation \cite{2024-Zhang-FewFedPIT-,2024-Toshniwal-OpenMathInstruct1-}, we investigate a FedDIT setup employing a server-hosted public dataset (a cross-domain instruction set, detailed in \cref{subsec:train_and_test_dataset_information}) and utilize various sampling strategies to realize the data augmentation (discussed in \cref{sec:preliminaries}). This dataset, abstracting diverse open-source instructions, provides a versatile foundation for augmentation techniques designed to elevate model performance within specific target domains.

Additionally, the factors affecting FedDIT are still unclear. Compounding this uncertainty, introducing augmented instructions may further complicate results, making it difficult to ascertain effective improvement strategies. We conduct experiments to unveil a significant finding: { there is no monotonic correlation between the degree of non-independent and identically distributed (non-iid) and LLM's performance in the context of FedDIT}. Inspired by Explore-Instruct \cite{2023-Wan-ExploreInstruct-P2CEMNLP}, which highlights the potential of domain coverage in domain-specific instruction tuning. Unlike previous metric that measures domain coverage using the distribution of verb-noun pairs, we define a more general, representation-based cross-client domain coverage metric and investigate its impact on FedDIT. Our results show that domain coverage significantly influences model performance within the corresponding domain.

To maximize cross-client domain coverage without compromising client data privacy, we propose a novel FedDIT algorithm, \textbf{FedDCA} (\textbf{D}omain \textbf{C}overage \textbf{A}ugmentation). The inability of direct data sharing under privacy constraints (detailed in \cref{subsec:optimization_problem}) motivates our approach. Initially, each client computes a set of candidate cluster centers from its local data; these serve as privacy-preserving proxies for its semantic distribution. FedDCA then employs a server-side, domain-coverage-oriented selection algorithm (detailed in \cref{subsec:greedy}) to choose a strategic subset (denoted as client centers) from all uploaded candidate centers. These client centers then guide server-side dense retrieval for instruction augmentation. This selection process serves two critical purposes: 1) it provides a privacy-preserving way to capture the semantic diversity of distributed data, and 2) it enables efficient retrieval of relevant public instructions that complement the collective needs of the clients. By strategically optimizing the cross-client domain coverage, FedDCA efficiently constructs an augmented train set that enhances the model's generalization capability, leading to superior performance on domain-specific tasks.

To substantiate our claims, we conduct a rigorous empirical study. This includes foundational experiments identifying the key performance drivers in FedDIT, the development and evaluation of our proposed FedDCA algorithm across multiple specialized domains (medical, financial, and mathematical) against eleven baselines, and in-depth analyses of its practical aspects such as privacy, scalability, and robustness in challenging settings (detailed in \cref{sec:experiments,sec:appendix_experiments}). Our key contributions are as follows:

\begin{itemize}[itemsep=2pt,topsep=0pt,parsep=0pt]
    \item We empirically reveal a critical finding: in Federated Domain-specific Instruction Tuning (FedDIT), cross-client domain coverage, rather than data heterogeneity, substantially impacts LLM effectiveness. We propose a novel representation-based metric to quantify this cross-client domain coverage.
    \item We propose FedDCA, an algorithm that maximizes cross-client domain coverage through diversity-oriented client center selection and retrieval-based instruction augmentation.
    \item Extensive experiments across diverse domains demonstrate FedDCA's superior effectiveness and plug-and-play capability, outperforming baselines by at most 29.19\% in model performance and improving relative domain coverage by 4.82\% to 21.36\%. We also show its privacy-preserving capability in \cref{subsec:privacy_analysis}.
    \item We demonstrate FedDCA's robustness and practical applicability by evaluating its performance in data selection scenarios (achieves comparable performance using only 10\% of baseline's data), held-out settings where the server lack of task-specific public data, and its consistent effectiveness across varying inter-client data heterogeneity (\cref{subsec:performance_analysis,sec:appendix_experiments,sec:robustness_heterogeneity}).
\end{itemize}

\section{Related Work} \label{sec:related_work}

\paragraph{Federated Instruction Tuning.}\label{para:FedIT} Instruction tuning has been widely applied across various application areas of large language models (LLM), serving as a key technique to enhance the capabilities and controllability of LLM \cite{2024-Zhang-Instruction-,2022-Wei-Finetuned-}. Recently, federated instruction tuning (FedIT) has emerged as an effective strategy for the distributed optimization of LLMs, leveraging federated learning (FL) protocols to improve the handling of privacy-sensitive tasks in real-world scenarios. So far, several FedIT frameworks \cite{2024-Ye-OpenFedLLM-,2024-Ye-FedLLMBench-,2024-Zhang-Building-} have been established to evaluate the effectiveness of FedIT across multiple datasets, tasks, and FL methods. While these platforms provide a foundation for research, they have not yet introduced more complex federated algorithms and deeply investigated the challenging problems and factors affecting FedIT, which are crucial for advancing this field.

Some progress has been made in the study of FedDIT \cite{2024-Zhang-FewFedPIT-,Wang2024KnowledgeSGPS,Ye2024LeveragingUT}, such as FewFedPIT, which addresses data scarcity by locally generating data using pre-trained LLMs and is the first to explore memory extraction attacks within FedIT. Another relevant approach is FedIT-U2S \cite{Ye2024LeveragingUT}, which focuses on enabling FedIT when clients only possess raw, unstructured documents. Similar to our conceptualization of a server-hosted public dataset for augmentation, FedIT-U2S also leverages server-side resources (the example database and the LLM for instruction generation) to enrich client data, albeit with a focus on structuring initially unstructured data rather than augmenting existing instructions with diverse new ones.

Non-IID data distribution is a common challenge in FL \cite{pmlr-v119-karimireddy20a,2020-Li-Federated-PMLS,sattler2019robust,li2019convergence,moon}. However, the impact of data heterogeneity on FedDIT has not been fully explored. To fully understand the impact of data heterogeneity on FedDIT and what truly matters in FedDIT, we conduct a comprehensive analysis in \cref{subsec:what_matter}.

\paragraph{Domain Instruction Augmentation.}\label{para:domain_instruction_augmentation} In the real world, there is an urgent need for training LLMs with specific functionalities (e.g., reasoning capabilities) or domain-specific LLMs (e.g., code \cite{2023-Nijkamp-CodeGen2-,2023-Luo-WizardCoder-}, medical \cite{2024-Zhang-AlpaCare-}, financial \cite{2023-Yang-InvestLM-,2023-Yang-FinGPT-,2023-Zhang-InstructFinGPT-,2023-Wu-BloombergGPT-}, mathematical \cite{2024-Yue-MAmmoTH2-,2023-Luo-WizardMath-}). 

Existing works tend to use open-source domain-specific instruction tuning datasets for training. However, the target domain may not always have corresponding ready-made domain-specific instruction datasets. Even if they exist, these datasets are often limited in scale. 

Augmentation strategies broadly fall into two categories: those leveraging existing/mined data and those generating new instructions. Methods based on \textbf{existing or mined data} include reusing human-curated public datasets \cite{2023-Wang-InstructUIE-,2023-Zhang-MultiTask-}, retrieving from large instruction pools \cite{2024-Xia-LESS-, 2023-Jiao-ParroT-}, or scaling instruction acquisition from the web \cite{2024-Yue-MAmmoTH2-, 2024-Zhou-JiuZhang30-}. These approaches offer access to diverse, potentially high-quality instructions with generally favorable efficiency and privacy, but depend on the availability and relevance of such external data for the target domain. Conversely, \textbf{generative methods} \cite{2023-Wang-SelfInstruct-,2024-Zhang-FewFedPIT-,Wang2024KnowledgeSGPS} create new instructions using LLMs (locally or via APIs). While potentially highly tailored, they often incur significant computational/API costs, raise privacy concerns, and face challenges in ensuring consistent instruction quality and diversity, as detailed in \cref{sec:appendix_experiments}.

In summary, adapting existing instruction augmentation techniques to federated learning presents significant challenges in balancing data quality, diversity, efficiency, privacy, and scalability. Current methods often involve trade-offs between costly, quality-variable generative approaches and data-dependent retrieval strategies. This necessitates robust augmentation frameworks tailored for FedDIT, motivating our work in this area.

\section{Problem Formulation}\label{sec:preliminaries}

Federated Domain-specific Instruction Tuning (FedDIT) is a federated learning approach designed to improve the performance of LLMs in specific domains by utilizing limited cross-client private data in combination with domain-specific instruction augmentation strategies \cite{2024-Zhang-FewFedPIT-,2024-Xia-LESS-,2023-Wang-SelfInstruct-}. 

For practical and scalable augmentation within FedDIT, we conceptualize the primary source of augmentation instructions as a \textbf{server-hosted, multi-domain public dataset} (detailed in \cref{subsec:train_and_test_dataset_information}). This strategic choice offers several key advantages in a federated context. Firstly, it provides a unified and manageable resource pool, abstracting the complex origins of public instructions (whether curated, mined, or pre-generated). Secondly, server management of this dataset decouples data sourcing from the core federated augmentation logic (such as retrieval strategies), allowing the latter to focus on optimizing instruction selection and distribution. Finally, a centralized public dataset enables efficient server-side pre-processing and indexing, which ensures consistent data quality and supports sophisticated retrieval mechanisms beneficial to all participating clients.

Consider $N$ distributed clients, each with local private data $D_k^l$ of size $N_k^l$, and augmented data $D_k^g$ from the server's public dataset $D^p$. Due to constraints like memory and computation, client $c_k$ can accept at most $N_k^p$ public instructions. The server maintains $D^p$ that spans multiple domains and is responsible for data augmentation strategies $\Lambda$ and parameter aggregation. For efficiency, we adopt Low-Rank Adaption (LoRA) \cite{2021-Hu-LoRA-} as the fine-tuning method, tuning additional parameters $\Delta\phi$ while keeping pre-trained LLM parameters $\phi$ frozen.

The objective of FedDIT is to enhance the domain-specific performance of LLMs through FL without sharing private data \cite{2024-Ye-OpenFedLLM-,2024-Zhang-FewFedPIT-,2024-Zhang-Building-}, which is defined as:

\begin{equation}\label{eq:global_objective}
    \setlength\abovedisplayskip{3pt}
    \setlength\belowdisplayskip{3pt}
    \argmin_{\Delta\phi} \{ F(\phi, \Delta\phi) \triangleq \sum_{k=1}^{N} p_k F_k \left(\phi, \Delta\phi_k; D_k; \alpha_k \right) \},
\end{equation}

where $F_k(\cdot)$ represents the accumulated instruction fine-tuning loss of model $w_{\phi+\Delta\phi_k}$ evaluated on client $c_k$'s augmented dataset $D_k^l \cup D_k^g$. Here, $p_k$ denotes client $k$'s weight based on data ratio, and $\alpha_k=\frac{N_k^p}{N_k^l+N_k^p}$ represents the proportion of public data. Equation \ref{eq:global_objective} optimizes the global model by minimizing the weighted sum of empirical losses across clients' augmented instructions, thereby enhancing in-domain utility. The empirical loss for client $c_k$, $F_k(\phi, \Delta\phi_k; \mathcal{D})$, is computed as $\frac{1}{|\mathcal{D}|}\sum_{j=1}^{|\mathcal{D}|}l(w_{\phi+\Delta\phi_k};x_j)$, where $x_j\in D, \forall j \in \{1,2,\dots, |D|\}$ and $l(\cdot)$ is the instruction tuning loss function.

For a client \(c_k\) with private instruction set \(\mathcal{D}_k\), we first encode instructions into embeddings using encoder \(w_{\mathrm{enc}}\). These \(d\)-dimensional vectors are then clustered into \(\xi\) groups via \(k\)-means, yielding centroid set $\mathcal{C}_k = \{\mathbf{c}_{k,1}, \mathbf{c}_{k,2}, \ldots, \mathbf{c}_{k,\xi}\}$, which is subsequently transmitted to the server.

We first introduce the \textbf{Direct Retrieval} baseline, which performs independent dense retrieval \cite{Zhao2022DenseTR} for each client without explicit consideration of cross-client domain coverage. For each centroid $\mathbf{c}_{k,j}$, the server retrieves the top-\(\frac{N_k^p}{\xi}\) most similar public instructions using cosine similarity. The retrieved instructions are aggregated as client $c_k$'s augmented data $D_k^p = \bigcup_{j=1}^{\xi}\mathcal{R}_{k,j}$. Finally, client $c_k$ fine-tunes its model \(w_k\) on the combined dataset $D^k = D_k^l \cup D_k^g$ of private and retrieved public instructions.

\section{What Truly Counts in FedDIT}\label{subsec:what_matter}

\begin{figure}[!tb]
\centering
\subfigure[Exp1. Med]{
    \begin{minipage}[b]{0.14\textwidth}
        \includegraphics[width=\textwidth]{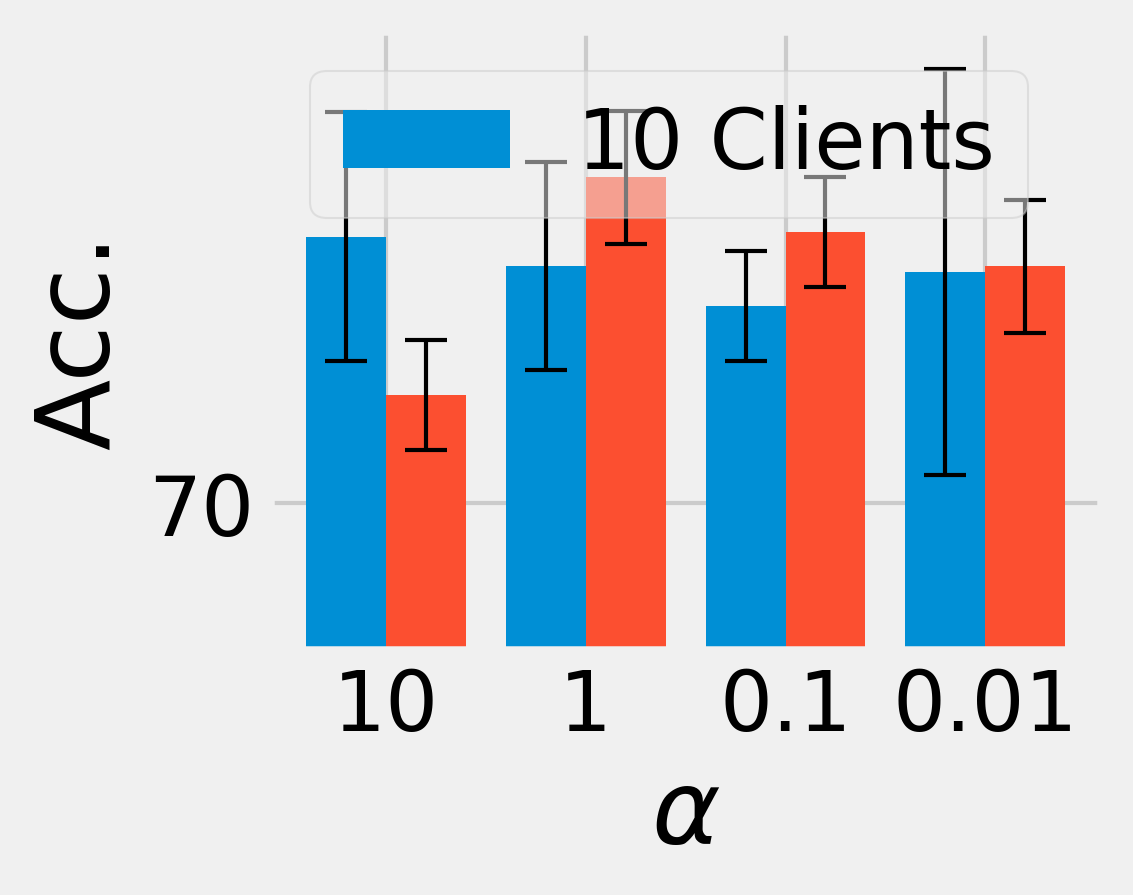}
    \end{minipage}
    \label{subfig:dirichlet_med}
}
\vspace{-1mm}
\hspace{-2mm}
\subfigure[Exp1. Fin]{
    \begin{minipage}[b]{0.14\textwidth}
        \includegraphics[width=\textwidth]{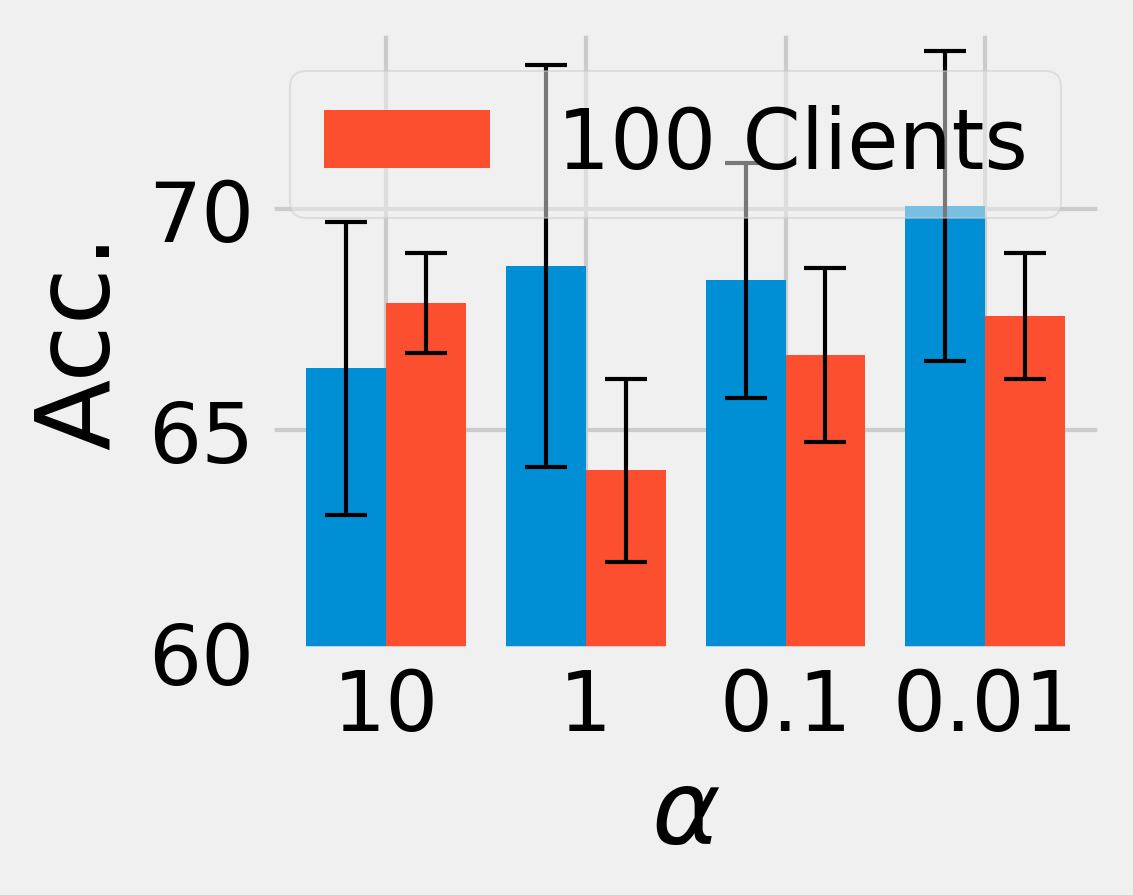}
    \end{minipage}
    \label{subfig:dirichlet_fin}
}
\vspace{-1mm}
\hspace{-2mm}
\subfigure[Exp1. Math]{
    \begin{minipage}[b]{0.14\textwidth}
        \includegraphics[width=\textwidth]{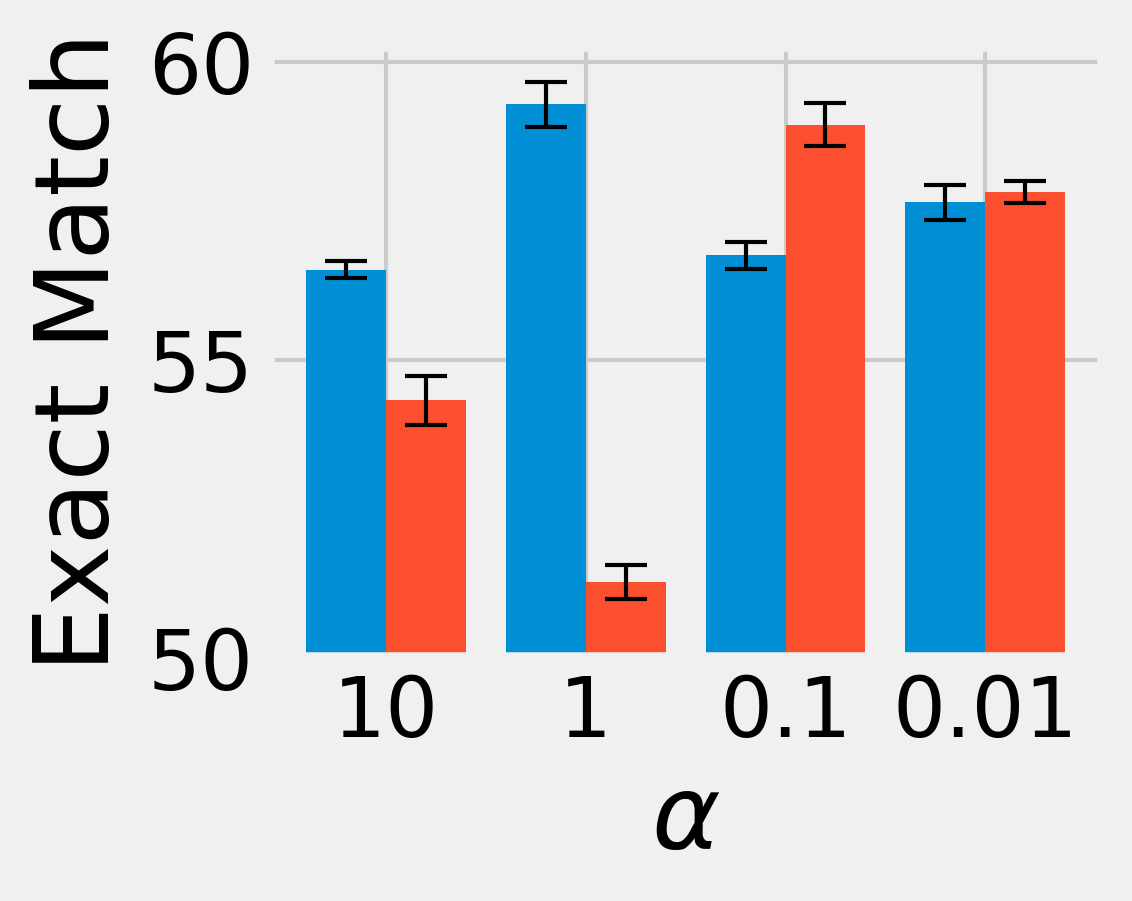}
    \end{minipage}
    \label{subfig:dirichlet_math}
}
\subfigure[Exp2. Performance (\%)]{
    \begin{minipage}[b]{0.15\textwidth}
        \includegraphics[width=\textwidth]{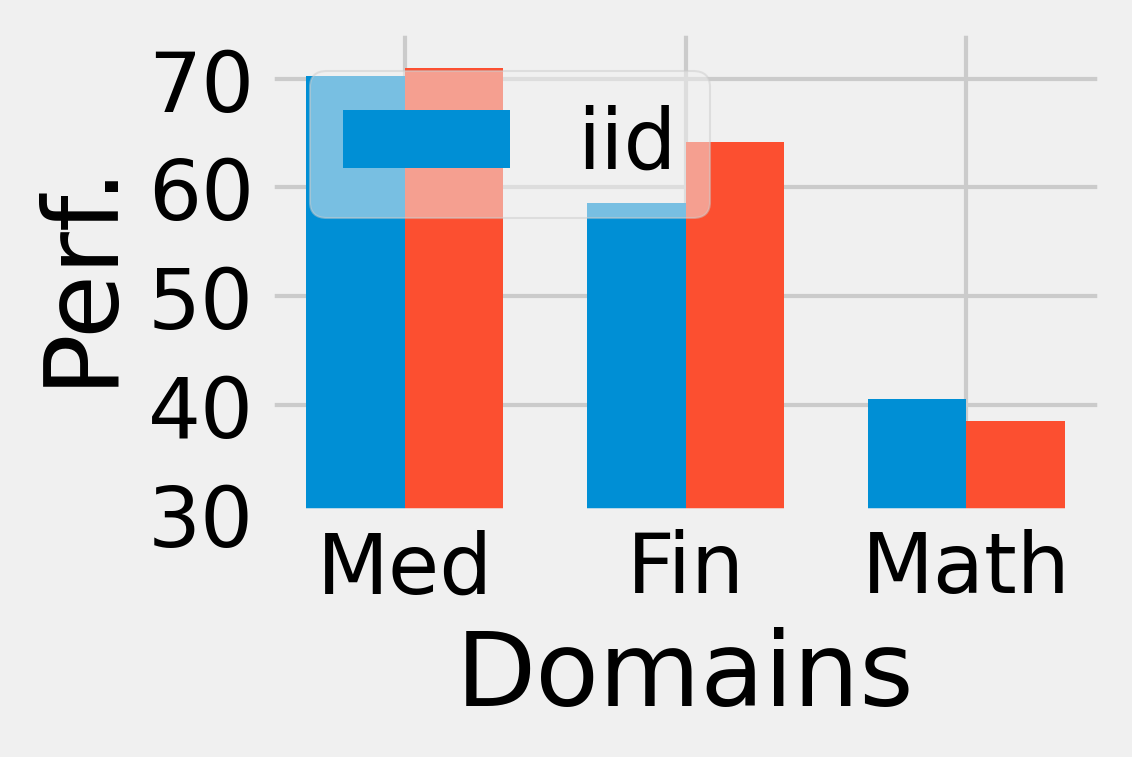}
    \end{minipage}
    \label{subfig:performance}
}
\hspace{2mm}
\subfigure[Exp2. Domain Coverage]{
    \begin{minipage}[b]{0.15\textwidth}
        \includegraphics[width=\textwidth]{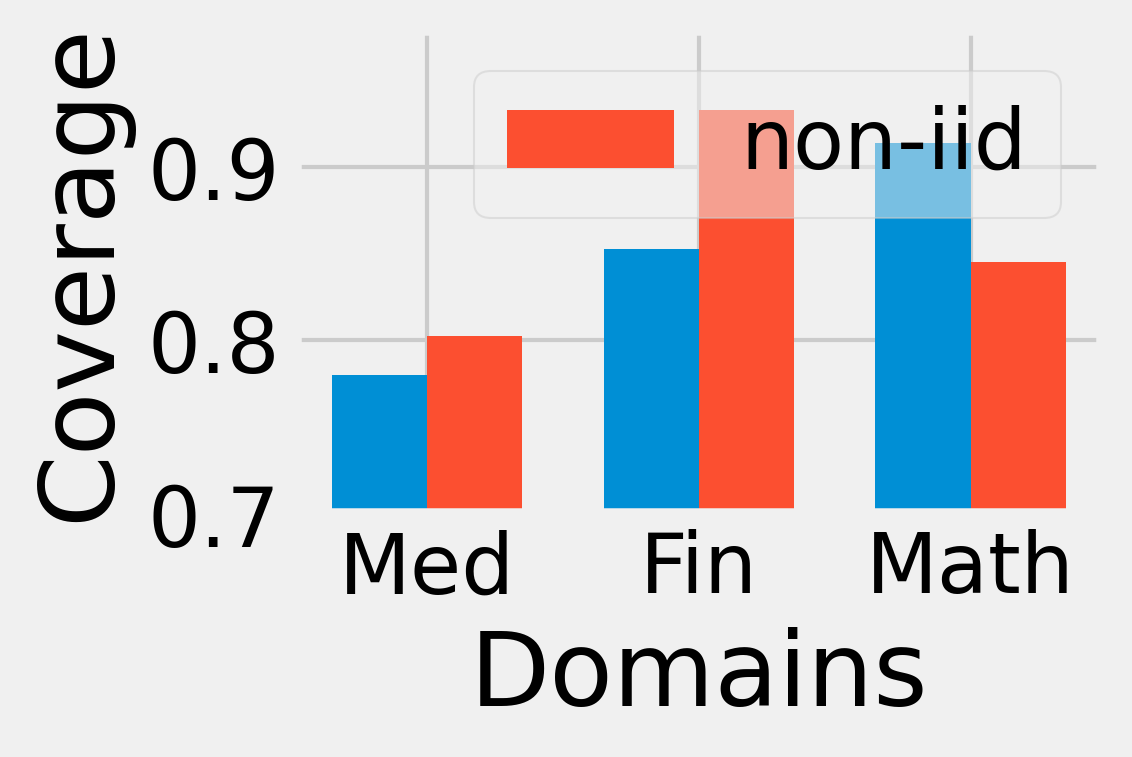}
    \end{minipage}
    \label{subfig:coverage}
}
\vspace{-2mm}
\caption{(a)-(c) Performance (\%) of different heterogeneity in each domain with 10 and 100 clients. (d)-(e) Performance and domain coverage of iid and non-iid settings on different domains. Experiments show that cross-client domain coverage significantly impacts model effectiveness, while data heterogeneity shows no monotonic correlation with performance.}
\label{fig:heterogeneity}
\end{figure}

\subsection{Data Heterogeneity: A Minor Factor}\label{subsubsec:not_matter}
We adopt the Dirichlet distribution  \cite{2020-Wang-Federated-,2019-Yurochkin-Bayesian-} to construct various heterogeneity ($\beta=[0.01, 0.1, 1, 10]$) and use k-means with $\xi=100$ to pseudo labeling instructions.

We then perform instruction tuning on both 10 and 100 clients with 2 randomly selected clients participating in each round. For each domain, we only use the in-domain data and then perform FedIT. We repeat the experiments for 3 times with different random seeds (42, 43 and 44) and report the average performance and the standard deviation in each domain. As shown in \cref{subfig:dirichlet_med,subfig:dirichlet_fin,subfig:dirichlet_math}, the performance of LLM shows a non-monotonic correlation with the data heterogeneity, which indicates that the performance of LLM does not directly depend on data heterogeneity and exist other factors playing a key role.

\subsection{Domain Coverage: A Key Factor}\label{subsubsec:matter}

Different from Explore-Instruct, which defines domain coverage through the distribution of verb-noun pairs, we attempt to conduct more in-depth and extensive experiments to study the effect of domain coverage on FedDIT. Firstly, we define the domain coverage in the FL setting, considering the cross-client data distribution. Assume the dataset of in-domain data $D^d$ represents the latent data distribution of this domain and the cross-client data is defined as $D^c = \cup_{k=1}^N \left(D_k^l \cup D_k^g\right)$.

Inspired by the submodular function \cite{krause2014submodular}, we first define a general coverage metric. For any two sets of item embeddings, $S_1$ (the reference set) and $S_2$ (the covering set), their coverage, denoted $d(S_1, S_2)$, is given by:
\begin{equation}\label{eq:domain_coverage}
    \setlength\abovedisplayskip{3pt}
\setlength\belowdisplayskip{3pt}
d(S_1, S_2) = \frac{1}{|S_1|} \sum_{s_1 \in S_1} \max_{s_2 \in S_2} \operatorname{sim}(s_1, s_2),
\end{equation}

where $\operatorname{sim}(\cdot,\cdot)$ is the cosine similarity between embeddings. Then the domain coverage of $D^c$ 
respect to $D^d$ is $d(D^d, D^c \cap D^d)$. The use of $D^c \cap D^d$ ensures that we only consider the in-domain portion of the aggregated client data for this particular coverage calculation, preventing misleading scores from out-of-domain client data, consistent with our original formulation.

To better align with our setup, we explore instruction augmentation based on both iid and non-iid cross-client local data distribution and then perform dense retrieval \cite{Zhao2022DenseTR} for FedDIT (the Direct Retrieval method). We set the number of clients to 10, while each client has 100 local instructions and obtains 1000 augmented public instructions from the server. For the iid setting, we randomly sample 1000 from the public dataset and divide them into 10 shards as each client's local data. For the non-iid distribution, we perform k-means clustering with $\xi=100$. Each client randomly samples 100 instructions from randomly selected distinct clusters and then performs retrieval for both settings. 

\cref{subfig:performance,subfig:coverage} presents the performance and the according domain coverage of FedDIT in different domains with iid and non-iid settings. We can observe that both iid and non-iid settings outperform in some domains, but both collectively indicate that higher domain coverage correlates with better performance.

\vspace{-2mm}
\section{Method} \label{sec:method}
\vspace{-2mm}

\begin{figure}[!tb]
\centering
\includegraphics[width=0.45\textwidth]{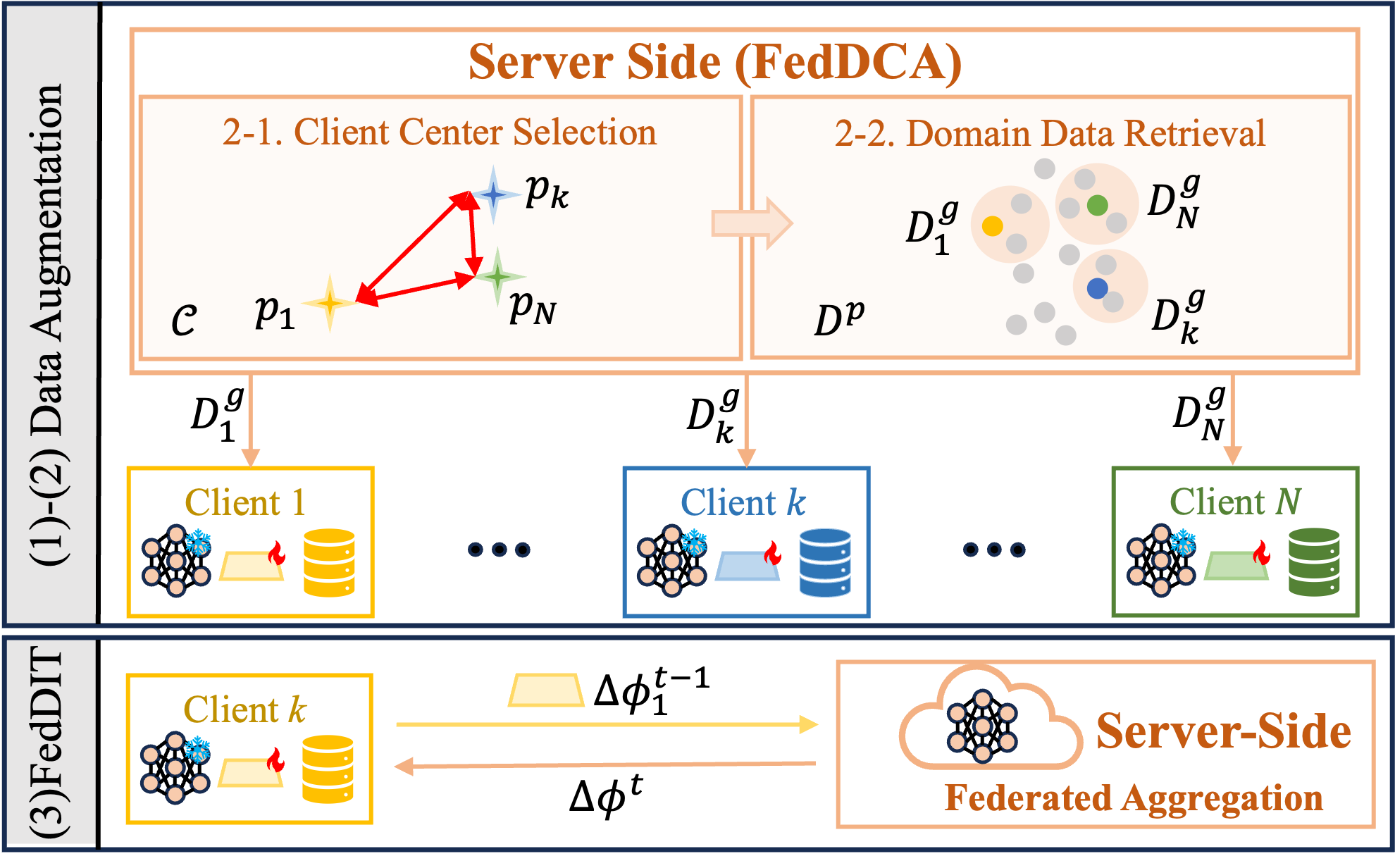}
\vspace{-2mm}

\caption{Overview, a three-stage process. \textbf{1) Client-side}: Clients $c_k$ perform local instructions clustering and send cluster centers $\mathcal{C}_k$ to the server. \textbf{2) Server-side Augmentation:} The server executes diversity-oriented client center selection to maximize domain coverage (obtaining $\mathcal{P}$), retrieves augmented data $D_k^g$ based on these centers, and distributes it to clients. \textbf{3) Collaborative Fine-tuning:} Clients collaboratively fine-tune the LLM, exchanging LoRA parameters $\Delta\phi$ with the server.}
\label{fig:overview}
\end{figure}

Based on the above empirical observations, we propose FedDCA, which enhances domain coverage to obtain a LLM that performs well on domain-specific tasks (shown in \cref{fig:overview}). We formulate the cross-client domain coverage optimization problem and then introduce the FedDCA algorithm.

\subsection{Optimization Problem}\label{subsec:optimization_problem}

As domain coverage directly affects the in-domain performance of the LLM, FedDCA aims to maximize the domain coverage of the cross-client augmented data $D^c$ with respect to the in-domain data distribution $D^d$. However, privacy constraints prevent clients from sharing local data, meaning the server lacks direct knowledge of clients' domain information. Consequently, directly identifying a cross-client dataset that maximizes domain coverage is infeasible. Instead, we leverage client-provided cluster centers ($\xi$ clusters per client, obtained by k-means algorithm \cite{2023-Vardakas-Global-}) as privacy-preserving proxies for local data. These centers, typically vector averages, effectively represent local distributions with manageable privacy risks (elaborated in \cref{subsec:privacy_analysis}). The core challenge thus shifts to selecting an optimal set of these client centers, $\mathcal{P}$ (one per client), such that retrieving public instructions based on $\mathcal{P}$ enhances cross-client domain coverage. This formulation provides a tractable approximation to the original optimization problem under inherent privacy constraints.

Additionally, in the FL setting, communication cost is always a critical factor. Thus, we formulate the optimization problem as follows:

\vspace{-2mm}   
\begin{equation}
    \label{eq:optimization}
    \setlength\abovedisplayskip{3pt}
\setlength\belowdisplayskip{3pt}
    \argmin_{\mathcal{P}} \left\{ \eta(\mathcal{P}) - d(D^d, \mathcal{P}) \right\}, 
\end{equation}
where the first term $\eta(\mathcal{P}) = \sum_{i=1}^N |\mathcal{P}_i|$ represents the communication overhead of transmitting client centers. And the second term is the domain coverage of the selected client center set $\mathcal{P}$, defined in Eq.\ref{eq:domain_coverage}. 

For simplicity, we let $\xi$ be the constant $N$ (further discussed in \cref{para:impact_of_different_cluster_number}). 
To better fit the FL environment and enhance the computational efficiency, we propose FedDCA as follows.

\subsection{FedDCA}\label{subsec:greedy}

Optimizing Eq.~\ref{eq:optimization} to find the ideal client center set $\mathcal{P}$ presents significant practical hurdles.
First, an exhaustive search over all $C_{\xi N}^{N}$ candidate center combinations is computationally infeasible due to its exponential complexity.
Second, directly leveraging a server-hosted public dataset $D^p$ to guide client center selection for optimal domain coverage is often impractical. This stems from several factors:
1) privacy constraints limit the server's knowledge of individual client data distributions and their specific domain focuses;
2) the inherent ambiguity and hierarchical nature of `domains' (e.g., `medical' versus specific medical sub-tasks) complicate precise server-side domain coverage calculation, which is already costly over a potentially massive $D^p$;
3) the assumption of a relevant $D^p$ being available on the server may not hold, particularly if clients utilize external retrieval or local \emph{self-instruct} for augmentation.

These challenges underscore the need for a more pragmatic selection strategy. Therefore, we propose FedDCA. As detailed in Alg.~\ref{alg:greedy_center_selection}, FedDCA seeks a sub-optimal solution in polynomial time (discussed in \cref{subsec:discussions}) and operates on the server through two main stages: 1) a coverage-oriented client center selection mechanism, and 2) client-center-based dense retrieval for data augmentation.

\paragraph{Coverage-oriented Client Center Selection.} 
To address these challenges, we propose a novel client center selection strategy that, crucially, \emph{does not rely on a server-side public dataset $D^p$}. Let $\mathcal{C}_{\text{all}} = \bigcup_{i=1}^N \mathcal{C}_i$ be the set of all candidate cluster centers uploaded by the $N$ clients (where each client $c_i$ provides $\xi$ centers in $\mathcal{C}_i$). Our approach leverages $\mathcal{C}_{\text{all}}$ in two ways: 1) We use $\mathcal{C}_{\text{all}}$ as a proxy for the target domain distribution. Domain coverage is then approximated by evaluating how well a selected subset of centers covers $\mathcal{C}_{\text{all}}$ itself, making the computation tractable and independent of $D^p$. 2) Instead of an exhaustive search, we employ an iterative, coordinate ascent-style algorithm (detailed in Alg.~\ref{alg:greedy_center_selection}) to select a final set $\mathcal{P}$ of $N$ client centers. This efficiently yields a high-quality, albeit potentially suboptimal $\mathcal{P}$. Based on the $\mathcal{C}_{\text{all}}$, the domain coverage could be calculated as $d(\mathcal{C}_{\text{all}}, \mathcal{P})$.



Specifically, each client \(c_i\) locally runs \(k\)-means to produce \(\xi\) candidate centers \(\mathcal{C}_i\), then uploads these centers to the server. The server initializes a selection \(\mathcal{P}\) by picking one cluster center from each client (thus \(|\mathcal{P}| = N\)) and computes the overall coverage solely based on these uploaded centers. In each iteration, FedDCA performs the following update (as shown in Alg.~\ref{alg:greedy_center_selection}): it removes the currently chosen center from \(\mathcal{P}\), tries replacing it with other candidate cluster centers in \(\mathcal{C}_{\text{all}}\), and keeps the replacement if it improves the cross-client domain coverage; this process is repeated until no further improvement is found. 

\begin{algorithm}[!tb]
\caption{FedDCA: Client Center Selection}
\label{alg:greedy_center_selection}
\setstretch{1}
\small
\begin{algorithmic}[1]
    \State \textbf{Initialize:} 
    \State \quad For each client \(c_i\), pick any initial center \(p_i \in \mathcal{C}_i\).
    \State \quad Let \(\mathcal{P} = \{p_1, \dots, p_N\}\) and compute its coverage \(C(\mathcal{P})\).
    \State \quad \(\Omega^* \gets C(\mathcal{P})\); \(i \gets 0\)
    \While{True}
        \State \(\mathcal{P}_{-i} \gets \mathcal{P} \setminus \{p_i\}\) 
            \Comment{Remove \(p_i\) from \(\mathcal{P}\)}
        \State \(\nu^* \gets p_i\); \(\Omega_{\text{old}} \gets \Omega^*\)

        \For{$q \in (\mathcal{C}_{\text{all}}-\mathcal{P}_{-i})$} 
            \State $\mathrm{C} = d(\mathcal{C}_{\text{all}}, \mathcal{P}_{-i} \cup \{q\})$ \Comment{Defined in Eq.~\ref{eq:domain_coverage}}
            \If {$\mathrm{C} > \Omega^*$}
                \State $\Omega^* \gets \mathrm{C}; \nu^* \gets q$
            \EndIf
        \EndFor
        
        \If{\(\Omega^* = \Omega_{\text{old}}\)}
            \State \textbf{break}
        \EndIf
        
        \State \(p_i \gets \nu^*\); \(\mathcal{P} \gets \mathcal{P}_{-i} \cup \{p_i\}\); \(C(\mathcal{P}) \gets \Omega^*\); \(i \gets (i + 1) \bmod N\)
    \EndWhile
    
    \State \textbf{Output:} The selected client center set \(\mathcal{P}\).
\end{algorithmic}
\end{algorithm}

This $D^p$-agnostic selection core is computationally tractable and versatile, effective for both data augmentation and targeted data selection through the selected client centers $\mathcal{P}$.

\paragraph{Domain Data Retrieval.} For each client center $p_k$, the server performs dense retrieval \cite{Zhao2022DenseTR} on public dataset $D^p$ to get the top-$N_k^p$ similar public instructions, then sends retrieved public datasets $\{D_1^g,\dots,D_N^g\}$ to each clients. 

Specifically, to avoid the overlap between public data and local private data, we set a threshold $\alpha$ to filter the public instructions that have a similarity score larger than $\alpha$ with the client center.

In summary, through the coverage-oriented client center selection and domain data retrieval, FedDCA obtains a better cross-client domain coverage of the augmented data $D_k^g$, which leads to a better model performance on the target domain. For the \textbf{Discussions} part please refer to \cref{subsec:discussions}.

\section{Experiments} \label{sec:experiments}
To demonstrate the effectiveness of FedDCA, we conduct extensive experiments across various domains and with several baselines. For additional results and analysis, please refer to \cref{sec:appendix_experiments}.

\begin{table*}[!tb]
    \centering
    \resizebox{0.9\textwidth}{!}{
    \begin{tabular}{@{\hspace{4pt}}l@{\hspace{4pt}}|@{\hspace{4pt}}c@{\hspace{4pt}}|@{\hspace{4pt}}c@{\hspace{4pt}}c@{\hspace{4pt}}c@{\hspace{4pt}}|@{\hspace{4pt}}c@{\hspace{4pt}}}
    \toprule
    Method & MMLU-Med & FPB & FiQA & TFNS & GSM8K \\
    \midrule
    Zero-shot & 70.60/- & 55.94/- & 18.54/- & 59.21/- & 23.27/- \\
    FedAvg \cite{2017-McMahan-Communicationefficient-AIS} & \underline{68.40}/0.6990 & 58.25/0.8529 & \underline{14.18}/0.8529 & 66.62/0.8529 & 47.46/0.7871 \\
    FedProx \cite{2020-Li-Federated-PMLS} & \underline{69.10}/- & 56.51/- & \underline{14.90}/- & 66.45/- & 47.15/- \\
    SCAFFOLD \cite{pmlr-v119-karimireddy20a} & \underline{70.20}/- & 62.71/- & \underline{15.27}/- & 66.49/- & 49.27/- \\
    FedAvgM \cite{2019-Hsu-Measuring-} & \underline{64.70}/- & 68.14/- & 29.27/- & 70.32/- & 46.85/- \\
    \midrule
    Random Sampling & 71.30/0.7940 & 64.19/0.9196 & \underline{13.09}/0.9196 & 65.53/0.9196 & \underline{47.38}/0.8651 \\
    Direct Retrieval & 72.20/0.8830 & 66.31/0.9293 & 19.11/0.9293 & 67.62/0.9293 & 50.87/0.8967 \\
    LESS \cite{2024-Xia-LESS-} & 71.00/0.7737 & 60.56/0.8917 & 16.00/0.8917 & 61.14/0.8917 & 43.13/0.8352 \\
    FewFedPIT \cite{2024-Zhang-FewFedPIT-} & 68.50/0.8250 & 56.30/0.8780 & 14.20/0.8780 & 59.10/0.8780 & 42.30/0.8380 \\
    Self-Instruct \cite{2023-Wang-SelfInstruct-} & 71.90/0.8586 & 59.73/0.9015 & 20.67/0.9015 & 66.54/0.9015 & 50.79/0.8811 \\
    KnowledgeSG \cite{Wang2024KnowledgeSGPS} & 73.13/0.9171 & 66.00/0.9490 & 32.72/0.9490 & 71.10/0.9490 & 51.20/0.9088 \\
    \textbf{FedDCA+FedAvg} & \textbf{74.50}/0.9348 & 67.24/0.9815 & 35.27/0.9815 & 73.32/0.9815 & \textbf{52.46}/0.9320 \\
    \textbf{FedDCA+FedProx} & 72.40/- & \textbf{72.93}/- & \textbf{38.18}/- & \textbf{77.55}/- & 51.25/- \\
    \textbf{FedDCA+SCAFFOLD} & 73.20/- & 72.68/- & 33.09/- & 75.50/- & 50.26/- \\
    \textbf{FedDCA+FedAvgM} & 68.90/- & 71.45/- & 31.45/- & 72.52/- & 49.76/- \\
    \bottomrule
    \end{tabular}
    }
    \vspace{-2mm}
    \caption{ Performance (\%) and domain coverage of FedDCA and other eleven baselines. Due to limited local data or inappropriate data augmentation strategy, occurs performance degradation in the Unaugmented and Random Sampling settings (underlined values). Different FL strategy does not affect domain coverage. We can see that FedDCA outperforms all other baselines.}\label{tab:performance}
    \vspace{-4mm}
    \end{table*}

\subsection{Experimental Setup}\label{subsec:experiment_setup}


\paragraph{Dataset and Evaluation Metrics.} To evaluate the performance of FedDCA, we conduct experiments utilizing a server-hosted public dataset consisting of multiple domains' data (detailed in \cref{subsec:train_and_test_dataset_information}): Alpaca, MedAlpaca \cite{2024-Zhang-AlpaCare-}, FinGPT \cite{2023-Yang-FinGPT-}, and MathInstruct \cite{2024-Toshniwal-OpenMathInstruct1-}. Performance is evaluated on MMLU-Med (medical); FPB, FiQA, and TFNS (financial); and GSM8K (math). The former two domains' metric is Acc. and math's is Exact Match. We set the number of clients to 10 by default, while each client has 100 local instructions and obtains 1k augmented public instructions by default from the server. Specifically, following the construction of the client's local data heterogeneity in \cref{subsec:what_matter}, we set the $\beta$ to 0.1 by default. We also exhibits FedDCA's strong robustness to different levels of data heterogeneity in \cref{sec:robustness_heterogeneity}.

\paragraph{Baselines.} 
We compare FedDCA against a range of baseline methods, categorized into unaugmented and augmented approaches (further details are available in \cref{sec:appendix_experiments}). \textbf{Unaugmented Methods.} These include \textbf{Zero-shot inference}, where the LLM directly predicts without any fine-tuning, and \textbf{\texttt{FedIT}}. \texttt{FedIT} represents the application of standard federated learning to clients' local data, for which we employ four widely recognized FL algorithms: FedAvg \cite{2017-McMahan-Communicationefficient-AIS}, FedProx \cite{2020-Li-Federated-PMLS}, SCAFFOLD \cite{pmlr-v119-karimireddy20a}, and FedAvgM \cite{2019-Hsu-Measuring-}. \textbf{Augmented Methods.} This category encompasses several strategies. \textbf{Random Sampling} serves as a simple baseline, randomly selecting $N_k^p$ public data instances for each client. \textbf{Direct Retrieval}, a precursor to our method, performs dense retrieval from public data based on each client's local instructions independently, without optimizing for cross-client domain coverage. \textbf{LESS} \cite{2024-Xia-LESS-} utilizes gradients from a warmed-up LLM on both training and a required validation set for its similarity-based retrieval. For generation-based approaches, \textbf{Self-Instruct} \cite{2023-Wang-SelfInstruct-} generates new instructions by prompting a powerful external LLM (e.g., GPT-3.5). \textbf{FewFedPIT} \cite{2024-Zhang-FewFedPIT-} is a federated few-shot method where the global model is used for local in-context learning to synthesize data, aiming to improve both performance and privacy. Finally, \textbf{KnowledgeSG} \cite{Wang2024KnowledgeSGPS} employs a server-hosted expert model to generate and refine synthetic data, which clients then use to enhance their local models.

\subsection{Main Results}\label{subsec:performance_analysis}

\paragraph{Performance \& Domain Coverage.} \cref{tab:performance} showcases the performance and corresponding domain coverage of FedDCA against various baselines across three distinct domains. FedDCA consistently outperforms all eleven baselines, demonstrating substantial performance improvements of up to 29.19\%. This superior performance is strongly correlated with its ability to achieve the highest domain coverage, surpassing other methods by an average of 4.82\% to 21.36\%. 

While unaugmented methods naturally struggle due to limited local data and thus restricted domain coverage, many existing augmentation strategies also exhibit significant drawbacks. 1) Generation-based methods present a mixed picture: KnowledgeSG, leveraging a powerful server-hosted expert model, achieves strong results while FewFedPIT underperform even unaugmented baselines, potentially due to inconsistent synthetic data quality and diversity. Moreover, these approaches often entail significant resource demands: KnowledgeSG and FewFedPIT involve high computational costs (server-side and client-side), while Self-Instruct incurs monetary expenses for API calls with the privacy leakage. The potential lack of quality and diversity in synthetic instructions further constrains their effectiveness. 2) Random Sampling and Direct Retrieval, while less costly, lack a strategic mechanism to ensure comprehensive cross-client domain coverage. 3) LESS, while performs domain specific retrieval, requiring access to a validation set, and additional computation (warmup training).


In conclusion, FedDCA's strategic data augmentation substantially enhances domain coverage and model performance, efficiently circumventing the computational burden and inherent quality/redundancy challenges of data generation.

\paragraph{Plug-and-Play.} We demonstrate FedDCA's plug-and-play capability by combining it with other augmentation methods. After sampling 1k instructions per client through different baselines, FedDCA performs retrieval by the selected client center (200 per client). As shown in \cref{tab:plug_and_play_performance}, despite using only 20\% of the sampled data, FedDCA achieves comparable performance with baselines. This highlights FedDCA's efficient data selection capability based on the cross-client domain coverage, which is further discussed in \cref{subsec:data_selection}.

\begin{table}[!tb]
\rmfamily
\centering
\resizebox{0.45\textwidth}{!}{
\begin{tabular}{@{\hspace{4pt}}l@{\hspace{4pt}}|@{\hspace{4pt}}c@{\hspace{4pt}}c@{\hspace{4pt}}c@{\hspace{4pt}}}
\toprule
\textbf{+ FedDCA} & MMLU-Med & FPB & GSM8K \\
\midrule
Random Sampling & 71.80 (\textbf{+0.50}) & 63.92 (-0.27) & 47.82 (\textbf{+0.44}) \\
Direct Retrieval & 72.10 (-0.10) & 67.78 (\textbf{+1.47}) & 52.53 (\textbf{+1.66}) \\
LESS & 70.80 (-0.20) & 61.03 (\textbf{+0.47}) & 43.92 (\textbf{+0.79}) \\
Self-Instruct & 70.70 (-1.20) & 60.21 (\textbf{+0.48}) & 50.25 (-0.54) \\
\bottomrule
\end{tabular}
}
\vspace{-2mm}
\caption{Plug-and-Play. Performance (\%) of FedDCA with different data augmentation methods. FedDCA matches baseline's performance by just 10\% of its data through selection.}\label{tab:plug_and_play_performance} 
\end{table}

\subsection{Ablation Study}

\paragraph{Impact of Different Cluster Number.}\label{para:impact_of_different_cluster_number} The hyperparameter $\xi$ is the number of clusters in the k-means algorithm. The experiment is conducted on $\xi=[N,2N,4N,8N]$, where $N$ is the number of clients. We report the domain coverage of the augmented dataset via FedDCA with different $\xi$ on the three domains in \cref{tab:k_combined}. Results show that there is no best $\xi$ for all domains and $\xi=N$ is usually an acceptable choice.


\begin{figure}[!tb]
\centering
\subfigure[Performance]{
    \begin{minipage}[b]{0.22\textwidth}
        \includegraphics[width=\textwidth]{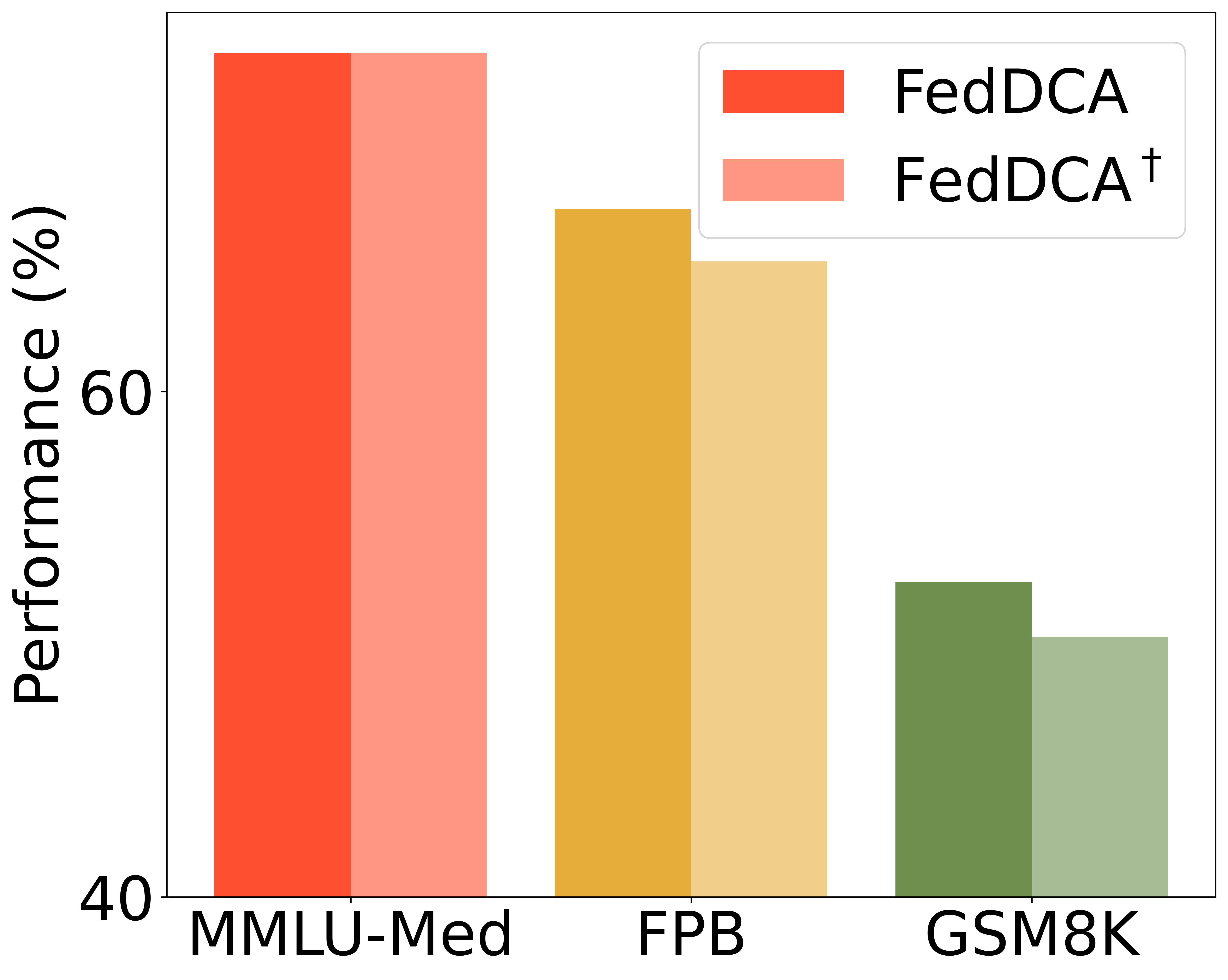}
    \end{minipage}
    \label{fig:threshold_performance}
}
\hspace{-2mm}
\subfigure[Domain Coverage]{
    \begin{minipage}[b]{0.22\textwidth}
        \includegraphics[width=\textwidth]{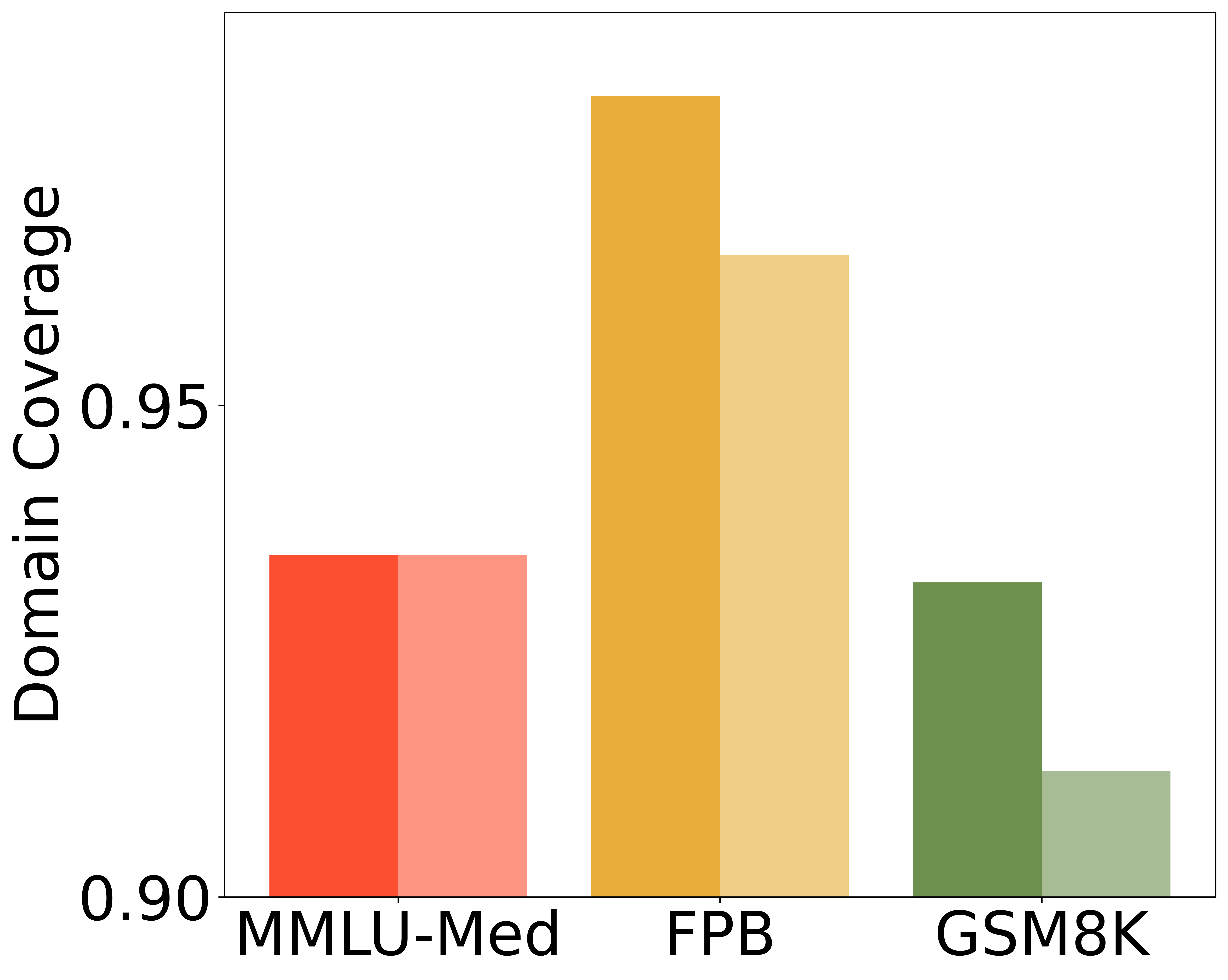}
    \end{minipage}
    \label{fig:threshold_coverage}
}
\vspace{-2mm}
\caption{Impact of similarity threshold $\alpha$ on FedDCA's (a) performance (\%) and (b) domain coverage. Results highlight the role of the similarity threshold in enhancing diversity. Lighter bars indicate FedDCA$^\dagger$.}
\label{fig:threshold}
\end{figure}



\paragraph{Similarity Threshold.} \cref{fig:threshold} presents an ablation study comparing FedDCA (w similarity threshold $\alpha$) against FedDCA$^\dagger$ (w/o $\alpha$). FedDCA achieves superior performance and domain coverage, highlighting the crucial role of the similarity threshold in effectively amplifying the diversity of augmented instructions and subsequently boosting model efficacy. Note that the medical domain remains unchanged, as the similarity of all pairwise embeddings is below 0.7 (the default value for the similarity threshold $\alpha$).

\begin{table}[!tbp]
\centering
\small
\resizebox{0.33\textwidth}{!}{
\begin{tabular}{@{\hspace{4pt}}l@{\hspace{4pt}}|@{\hspace{4pt}}c@{\hspace{4pt}}c@{\hspace{4pt}}c@{\hspace{4pt}}c@{\hspace{4pt}}}
\toprule
Domain & $\xi=10$ & $\xi=20$ & $\xi=40$ & $\xi=80$ \\
\midrule
Med. & 0.9348 & 0.9478 & 0.9466 & 0.9618 \\
Fin. & 0.9815 & 0.9814 & 0.9819 & 0.9813 \\
Math. & 0.9320 & 0.9348 & 0.9344 & 0.9337 \\
\bottomrule
\end{tabular}
}
\vspace{-2mm}
\caption{Domain Coverage for Different $\xi$ Values. $\xi=N$ is usually an acceptable choice.}
\label{tab:k_combined}
\end{table}

We also investigate the impact of different retrieval amounts on FedDCA's performance, detailed in \cref{subsec:retrieval_num}.


\section{Conclusion}


This work advances our understanding of Federated Domain-specific Instruction Tuning (FedDIT), establishing cross-client domain coverage as paramount over data heterogeneity for high performance. FedDCA provides a practical pathway to optimize this coverage, demonstrating significant gains through strategic data augmentation. FedDCA's plug-and-play capability and validated robustness in challenging scenarios (e.g., scarcity of task-specific public data, large-scale deployments) further positions it as a reliable building block for real-world, privacy-preserving specialized LLM applications. These findings offer both an applicable solution and crucial guidance for future federated system design where domain coverage is the key.

\section*{Limitations}

While FedDCA provides a robust and scalable framework for federated domain-specific instruction tuning, several avenues remain for further exploration. Although we have empirically demonstrated FedDCA's resilience to the similar cross-client data distributions (see \cref{sec:robustness_heterogeneity}), its performance may still be influenced by the quality and diversity of both the server-hosted public data and the client-provided candidate centers. Future work could investigate the adaptive client-side clustering to further enhance coverage and robustness.
Additionally, automating hyperparameter selection (e.g., the number of clusters per client) and extending FedDCA to handle evolving data distributions or continual learning scenarios are promising directions to further improve its practicality and generalization in real-world deployments.

\bibliography{custom}

\appendix

\section{Methodology}\label{sec:methodology}

\subsection{Discussions}\label{subsec:discussions}

\paragraph{Computation.}\label{para:computation} 
The client center selection process, as detailed in Alg.~\ref{alg:greedy_center_selection}, operates on $\mathcal{C}_{\text{all}}$, the set of all $N \cdot \xi$ candidate centers uploaded by $N$ clients (each providing $\xi$ centers).
Since FedDCA is an iterative optimization algorithm, let $T$ be the number of passes until convergence. In each pass, the algorithm tries to swap the currently selected center $p_i$ with one of the other candidate centers from $\mathcal{C}_{\text{all}}$ ($N\xi-N$ choices). Each potential swap requires re-evaluating the domain coverage $\mathrm{C}$, which is $N\xi(N+N\log N)$.
Therefore, the total complexity for one pass is approximately $O(N\xi \cdot (N\xi(N+N\log N))) = O(N^3\xi^2(1+\log N))$.
If the algorithm converges in $T$ passes, the total complexity for client center selection is $O(T \cdot N^3\xi^2(1+\log N))$. Overall, this complexity is polynomial in the number of clients $N$ ($\xi=N$ by default), rendering the selection process tractable for typical federated learning scenarios where $N$ and $\xi$ are manageable. 

To empirically validate the convergence speed ($T$) of our client center selection algorithm, we conducted the client center selection for each domain under the default setup for 100 times. The results show that FedDCA converges remarkably quickly, with an average of 1.1N passes across all experiments. This rapid convergence indicates that the algorithm typically requires only one full iteration through all clients, with occasional minor adjustments in a second round of iteration, making it highly efficient in practice.

\paragraph{Communication.}\label{para:communication}
The communication overhead of FedDCA is mainly incurred in two phases: (1) In the domain instruction augmentation phase, each client uploads $\xi$ cluster centers (typically low-dimensional vectors, e.g., 1024-d) to the server, and the server returns $N_k^p$ retrieved public instructions to each client. (2) During model parameter synchronization, FedDCA follows standard federated learning (FedIT) procedures, requiring only the exchange of LoRA parameters $\Delta\phi_k$, which significantly reduces communication cost.

\paragraph{Privacy.}\label{para:privacy} Comparing FedDCA with other FedIT methods \cite{2024-Zhang-FewFedPIT-,2024-Ye-OpenFedLLM-}, the difference lies in the client center selection stage. In this stage, the client only uploads the cluster center to the server, which is the average of embeddings to its cluster. In addition, the potential privacy leakage can be further avoided through homomorphic encryption \cite{2018-Acar-Survey-ACS}, which allows the server to directly compute on ciphertext for matrix multiplication for dense retrieval. 

\paragraph{Robustness to Outliers.} The k-means clustering, a component of our client-side processing, can be sensitive to outliers in the local data. If a client's local dataset contains significant outliers, these could potentially skew the resulting cluster centers, and subsequently affect the server-side selection and retrieval process. To mitigate this, two main approaches could be considered. \textbf{Firstly}, more robust clustering algorithms from existing literature could be adopted by clients. For instance, \citealp{Deshpande2023ImprovedORobust} enhances k-means++'s robustness by capping the selection probability for new centers, based on the overall clustering cost and an allowed number of outliers. This prevents distant, potentially outlier, points from dominating the initialization process and improves clustering in noisy data. \textbf{Secondly}, clients could perform preliminary data filtering or cleaning steps before initiating the FedDCA process. This might involve simple heuristic-based filtering (e.g., keyword-based removal of irrelevant instructions) or leveraging lightweight LLM-based tagging to identify and exclude potential outliers \cite{Bernsohn2024LegalLensLL,Biester2024LLMCleanCT,Zhang2024DataCUCocoon}. A comprehensive investigation into the impact of outliers and the optimal strategies for enhancing FedDCA's robustness in such scenarios is a promising direction for future work.

\subsection{Theoretical Foundation of FedDCA}\label{subsec:submodular_equivalence}

FedDCA's optimization objective is equivalent to the classic submodular facility location problem, ensuring our greedy algorithm achieves at least a $1-1/e$ approximation ratio.

\paragraph{The Optimization Problem in FedDCA.}
The overall optimization problem in our paper is formulated as Eq.~\ref{eq:optimization}. In our specific problem setting, the number of cluster centers ($\xi$) per client is considered fixed. Therefore, minimizing the objective function is equivalent to maximizing the domain coverage term $d(D^d, \mathcal{P})$ through the selected client centers $\mathcal{P}$.

\paragraph{The Submodular Facility Location Problem.}
In its abstract form, we are given: a set of ``locations'' or ``customers'' $\mathcal{U}$ that need to be serviced; a set of potential ``facilities'' $\mathcal{F}$ that can be opened; a benefit function $w(i, j)$ that quantifies the value of servicing customer $j \in \mathcal{U}$ with facility $i \in \mathcal{F}$. The objective is to select a subset of facilities $\mathcal{S} \subseteq \mathcal{F}$ of a given size $k$ to maximize the total benefit provided to all customers:
\begin{equation}
    g(\mathcal{S}) = \sum_{j \in \mathcal{U}} \max_{i \in \mathcal{S}} w(i, j)
\end{equation}

\paragraph{Formal Equivalence.}
The connection between our domain coverage problem and the facility location problem becomes clear when we map the components: \textbf{1) Customers ($\mathcal{U}$)} $\rightarrow$ The set of in-domain data, $D^d$; \textbf{2) Facilities ($\mathcal{F}$)} $\rightarrow$ The set of all candidate centers, $\mathcal{C}_{\text{all}}$; \textbf{3) Benefit ($w(i, j)$)} $\rightarrow$ The cosine similarity between two centers, $\operatorname{sim}(p, c)$.

Candidate centers are obtained by clustering each client's local data, whereas $D^d$ is the union of all public in-domain data points. This ``non-diagonal'' mapping still matches the standard facility-location structure because every data point is evaluated against the chosen center set, regardless of origin.

With this mapping, our domain coverage objective $d(D^d, \mathcal{P})$ has the form of the classic facility location function $g(\mathcal{P})$ normalized by the size of the in-domain data, $|D^d|$. That is, $d(D^d, \mathcal{P}) = \frac{1}{|D^d|}g(\mathcal{P})$.

\begin{table*}[!tb]
    \centering
    \small
    \begin{tabular}{@{\hspace{4pt}}l@{\hspace{4pt}}|@{\hspace{4pt}}c@{\hspace{4pt}}c@{\hspace{4pt}}c@{\hspace{4pt}}c@{\hspace{4pt}}c@{\hspace{4pt}}c@{\hspace{4pt}}}
    \toprule
    \multirow{2}{*}{\textbf{Method}} & \multirow{2}{*}{\makecell{Privacy\\Preserving}} & \multirow{2}{*}{\makecell{API\\Cost}} & \multirow{2}{*}{\makecell{Additional\\Information}} & \multirow{2}{*}{\makecell{Performance\\Degradation}} & \multirow{2}{*}{\makecell{Domain Coverage\\Oriented}} & \multirow{2}{*}{\makecell{Computational Overhead of \\ Data Augmentation}} \\
    & & & & & & \\
    \midrule
    FedAvg \cite{2017-McMahan-Communicationefficient-AIS} & {\color{green}\CheckmarkBold} & {\XSolidBrush} & {\XSolidBrush} & {\color{green}\CheckmarkBold} & {\XSolidBrush} & -\\
    Random Sampling & {\color{green}\CheckmarkBold} & {\XSolidBrush} & {\XSolidBrush} & {\color{green}\CheckmarkBold} & {\XSolidBrush} & -\\
    LESS \cite{2024-Xia-LESS-} & {\color{green}\CheckmarkBold} & {\XSolidBrush} & {\color{green}\CheckmarkBold} & {\color{green}\CheckmarkBold} & {\XSolidBrush} & high\\
    FewFedPIT \cite{2024-Zhang-FewFedPIT-} & {\color{green}\CheckmarkBold} & {\XSolidBrush} & {\XSolidBrush} & {\XSolidBrush} & {\XSolidBrush} & high\\
    KnowledgeSG \cite{Wang2024KnowledgeSGPS} & {\color{green}\CheckmarkBold} & {\XSolidBrush} & {\XSolidBrush} & {\XSolidBrush} & {\XSolidBrush} & high\\
    Self-Instruct \cite{2023-Wang-SelfInstruct-} & {\XSolidBrush} & {\color{green}\CheckmarkBold} & {\XSolidBrush} & {\XSolidBrush} & {\XSolidBrush} & low\\
    Direct Retrieval & {\color{green}\CheckmarkBold} & {\XSolidBrush} & {\XSolidBrush} & {\XSolidBrush} & {\XSolidBrush} & low\\
    \textbf{FedDCA (ours)} & {\color{green}\CheckmarkBold} & {\XSolidBrush} & {\XSolidBrush} & {\XSolidBrush} & {\color{green}\CheckmarkBold} & low\\
    \bottomrule
    \end{tabular}
    \vspace{-2mm}
    \caption{ Key differences between FedDCA and other baselines. FedDCA demonstrates the ability of: 1) privacy-preserving, 2) no API cost, 3) no additional information required, 4) avoiding performance degradation, and 5) aiming at domain coverage optimization. LESS requires additional information, as it needs access to the validation set for gradient-based retrieval.}
    \label{tab:comparison_methods}
    \end{table*}

\begin{table}[!tbp]
    \small
    \centering
    \resizebox{0.35\textwidth}{!}{
    \begin{tabular}{@{\hspace{4pt}}l@{\hspace{4pt}}|@{\hspace{4pt}}l@{\hspace{4pt}}l@{\hspace{4pt}}}
    \toprule
    Dataset & Size & Metric \\
    \midrule
    Alpaca & 52,002 & - \\
    MedAlpaca & 33,955 & - \\
    MMLU-Med$^\dagger$ & 1,089 & Acc. \\
    FinGPT & 76,772 & - \\
    FPB$^\dagger$ & 152 & Acc. \\
    FiQA$^\dagger$ & 35 & Acc. \\
    TFNS$^\dagger$ & 299 & Acc. \\
    MathInstruct & 224,567 & - \\
    GSM8K$^\dagger$ & 1,319 & Exact Match \\
    \bottomrule
    \end{tabular}
    }
    \caption{{Dataset information of each domain. Public dataset is composed of these five train sets (Alpaca, MedAlpaca, FinGPT, MathInstruct, and GSM8K). Test sets are marked with $\dagger$ and used for evaluation.}}
    \label{tab:dataset_info}
    \end{table}

\paragraph{Monotonicity and Submodularity.}
The facility location function $g(\mathcal{P})$ is known to be both monotone and submodular. Our objective $d(D^d, \mathcal{P})$ inherits these properties.

\textbf{Monotonicity}: This property is guaranteed because our benefit function, $\operatorname{sim}(p, c)$, is the cosine similarity from a sentence encoder, which is non-negative\footnote{\url{https://huggingface.co/BAAI/bge-large-en-v1.5}} (we have plotted histograms of the similarity scores for each domain, which fall within the range about [0.2, 0.85]). When adding a new center to $\mathcal{P}$, the maximum similarity for any data point $c$ can only increase or stay the same. If a negative value ever arises, we can rectify the score via applying an affine shift, which preserves the order of similarities and therefore the monotonicity. Consequently, the total sum of maximum similarities, $g(\mathcal{P})$, cannot decrease.

\textbf{Submodularity (Diminishing Returns)}: To formally show this, let's define the marginal gain for a single data point $c \in D^d$ when adding a new center $x$ to a set of centers $S$ as:
\begin{equation}
    \Delta_c(S,x)=\max\bigl(0, \operatorname{sim}(x,c)-\max_{p\in S}\operatorname{sim}(p,c)\bigr)
\end{equation}

This formula calculates the new coverage a center $x$ provides to a point $c$, beyond what is already provided by the existing set $S$. Now, consider two sets of centers $A$ and $B$ such that $A \subseteq B$. For any data point $c$, the maximum similarity from set $B$ must be at least as large as from set $A$, so:
\begin{equation}
    \max_{p\in A}\operatorname{sim}(p,c) \le \max_{p\in B}\operatorname{sim}(p,c)
\end{equation}
Consequently, the marginal gain from adding $x$ must be smaller for the larger set $B$. That is, for any $c$:
\begin{equation}
    \Delta_c(A,x) \ge \Delta_c(B,x)
\end{equation}

The total marginal gain of adding center $x$ to a set $S$ is:
\begin{equation}
    g(S \cup \{x\}) - g(S) = \sum_{c \in D^d} \Delta_c(S,x)
\end{equation}
Since the inequality holds for each individual term in the sum, it must hold for the sum itself. This demonstrates the diminishing returns property:
\begin{equation}
    g(A \cup \{x\}) - g(A) \ge g(B \cup \{x\}) - g(B) \text{ for } A \subseteq B
\end{equation}

Overall, our problem of maximizing the domain coverage $d(D^d, \mathcal{P})$ is equivalent to solving the facility location problem of maximizing $g(\mathcal{P})$, ensuring theoretical guarantees for FedDCA.

In addition, we have provided the time complexity analysis in \cref{subsec:discussions} and our empirical results show that the FedDCA algorithm converges quickly (about 1.1N passes). Furthermore, \cref{subsec:empirical_tightness} demonstrates that FedDCA achieves near-optimal performance in practice.

\section{Experiments}\label{sec:appendix_experiments}
\subsection{Train and Test Dataset Information}\label{subsec:train_and_test_dataset_information}
We evaluate our method on multiple domains including medical (MedAlpaca, MMLU-Med), financial (FinGPT, FPB, FiQA, TFNS), and mathematical reasoning (MathInstruct, GSM8K). Table~\ref{tab:dataset_info} shows the dataset statistics for each domain. The public dataset used for retrieval consists of the training sets from these domains and the general instruction dataset (Alpaca) through concatenation and random shuffling, while the test sets (marked with $\dagger$) are used for evaluation.

\subsection{Baselines}\label{subsec:baselines}

To highlight FedDCA's advantages, Table~\ref{tab:comparison_methods} provides a comparative overview with existing baseline methods along six critical dimensions: Privacy Preserving, API Cost, Additional Information requirements, potential for Performance Degradation, orientation towards Domain Coverage, and the Computational Overhead of Data Augmentation. These dimensions are crucial for assessing the practical viability of federated instruction tuning strategies. We analyze each dimension in detail below:

\begin{figure*}[!tbp]
    \centering
    \includegraphics[width=0.8\textwidth]{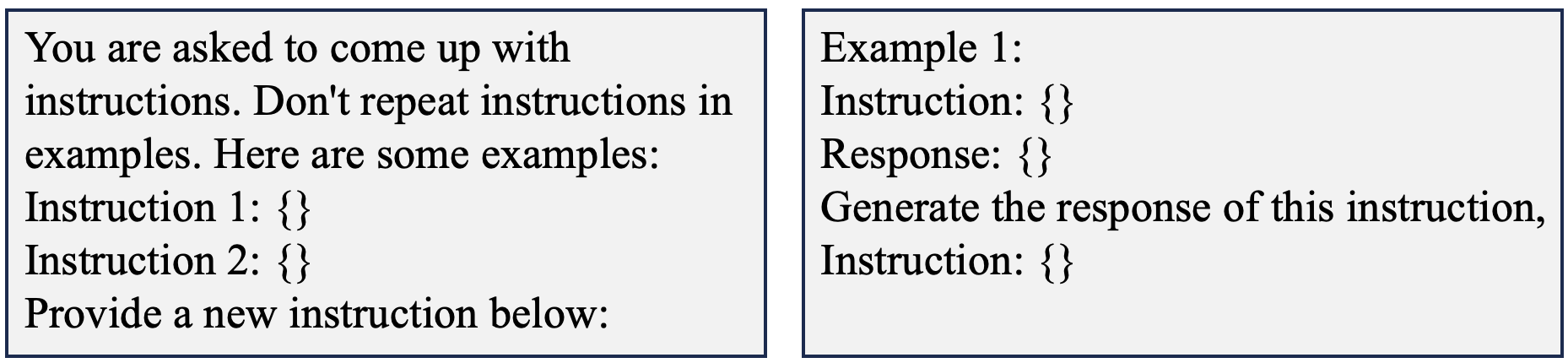}
    \caption{Prompts used in the Self-Instruct data generation. (a) Prompt for generating new instructions. Two examples are randomly sampled from the client's local data for in-context demonstration. (b) Prompt for generating responses. We prompt GPT-3.5 to generate responses with a randomly selected example for one-shot in-context learning.}
    \label{fig:instruction}
 \end{figure*}

\begin{figure}[!tbp]
\centering
\includegraphics[width=0.4\textwidth]{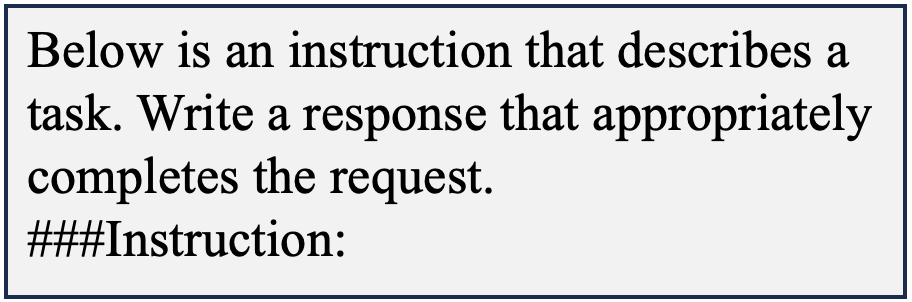}\caption{Prompt used for memory extraction attack.}
\label{fig:memory_extraction}
\end{figure}

\begin{itemize}[itemsep=2pt,topsep=0pt,parsep=0pt]
    \item \textbf{Privacy Preserving}: Most evaluated methods, including FedDCA, inherently preserve client data privacy by design, as they do not share raw local data. Self-Instruct is a notable exception as it involves sending prompts containing potentially sensitive information to an external API. Our detailed privacy analysis for FedDCA can be found in \cref{subsec:discussions,subsec:privacy_analysis}.
    
    \item \textbf{API Cost}: FedDCA operates without requiring external API calls, thus incurring no associated costs. In contrast, Self-Instruct, by leveraging large models like GPT for instruction generation, entails API expenses.
    
    \item \textbf{Additional Information}: FedDCA primarily relies on clients' local training data for its operations. Some methods, such as LESS \cite{2024-Xia-LESS-}, necessitate additional resources like a validation set for its gradient-based retrieval mechanism.
    
    \item \textbf{Performance Degradation Avoidance}: A key objective is to enhance model performance or, at minimum, avoid degradation. As indicated in our main results (Table~\ref{tab:performance}), FedDCA is designed to prevent performance drops. In contrast, methods like FedAvg (when unaugmented), Random Sampling, and even some generation-based approaches like FewFedPIT and KnowledgeSG (if generation quality is not optimal or well-aligned) can risk performance degradation.
    
    \item \textbf{Domain Coverage Oriented}: FedDCA is uniquely engineered to explicitly optimize cross-client domain coverage in distributed environments. This strategic focus is a distinguishing feature largely absent in the other compared baselines, which typically do not have a direct mechanism for maximizing instruction diversity across the federation.
    
    \item \textbf{Computational Overhead of Data Augmentation}: FedDCA maintains a low computational overhead for its data augmentation phase, comparable to methods like LESS (due to gradient computation and warm-up), FewFedPIT (local LLM generation), and KnowledgeSG (expert model generation on the server) inherently involve higher computational demands for data augmentation.
\end{itemize}

In conclusion, these comparisons underscore FedDCA's well-rounded design. It effectively addresses key practical challenges by preserving privacy, avoiding API costs, minimizing reliance on additional information, preventing performance degradation, strategically optimizing domain coverage, and maintaining a low computational overhead for augmentation.

\subsection{Implementation Details}\label{subsec:implementation_details}

We consider FedDIT in the cross-device scenario, $N$ = 10 clients, $\mathcal{R}$ = 30 rounds, where we randomly sample 2 clients to be available for each round. Then, each available client performs FedDIT for 10 steps with AdamW optimizer, and the batch size is $B=32$ in a round. The initial learning rate is $5e-5$ with a cosine learning rate scheduler. Our experiment utilizes the widely used LLM, Llama3-8B\footnote{\url{https://huggingface.co/meta-llama/Meta-Llama-3-8B}} as the base model with 2048 max sequence length and adopts LoRA tuning method. The rank of LoRA is 16, and the scalar alpha is 16. For k-means \cite{2023-Vardakas-Global-}, we set cluster num $\xi$ = 10 and for FedDCA we set the similarity threshold $\alpha$ = 0.7. 
We utilize \texttt{bge-large-en-v1.5}\footnote{\url{https://huggingface.co/BAAI/bge-large-en-v1.5}} as both the client and server's encoder as default, which outputs embeddings of 1024 dimensions.

\begin{figure}[!tb]
    \centering
    \includegraphics[width=0.4\textwidth]{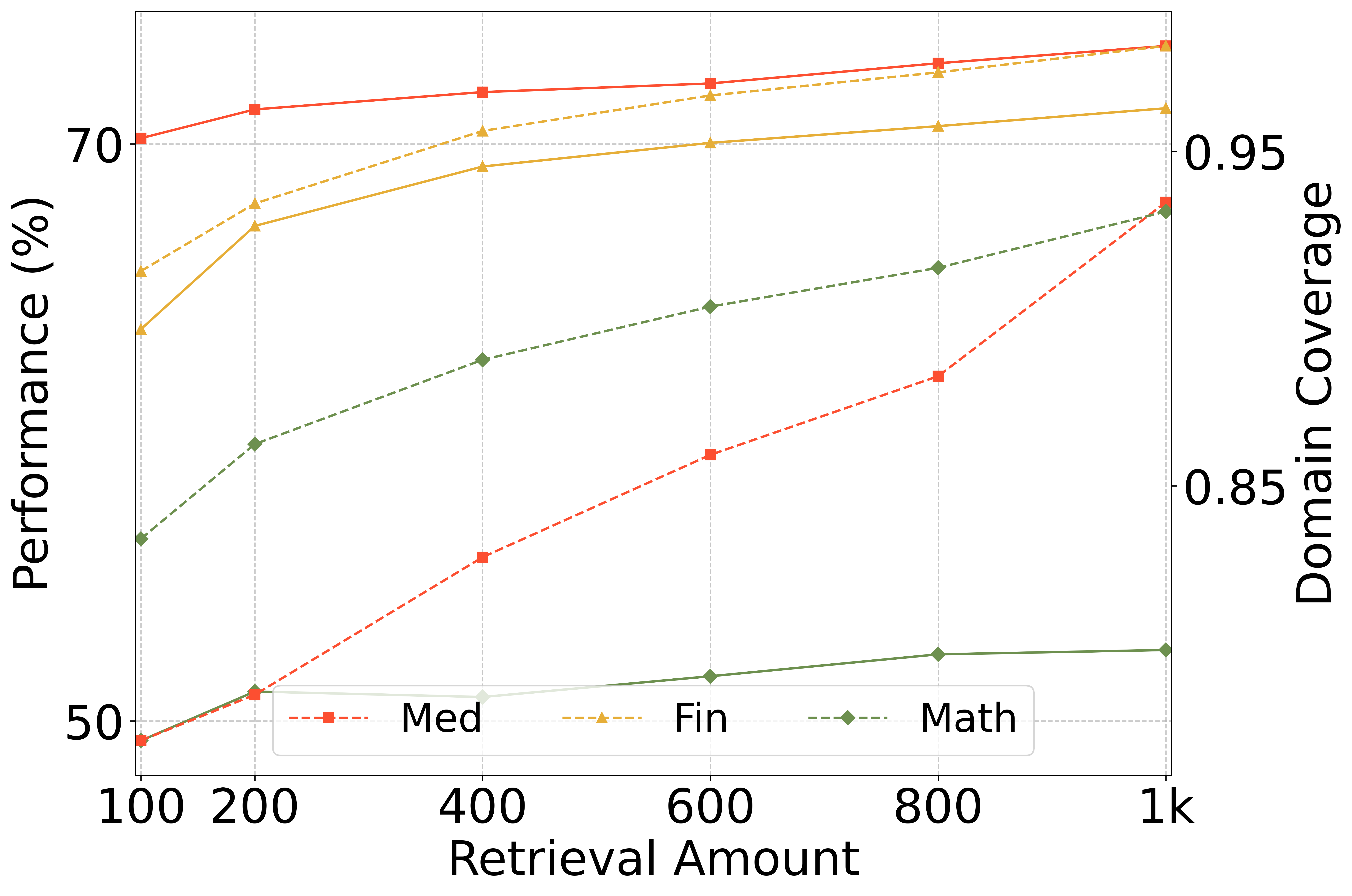}
    \vspace{-2mm}
    \caption{Impact of different retrieval amounts on FedDCA's performance (\%) and domain coverage. The dotted lines represent the domain coverge. Results demonstrate FedDCA's robust applicability across different augmentation scales.}
    \label{fig:effect_different_retrieval_numbers}
    \end{figure}

Specifically, for LESS, we uses LoRA with rank 128 and dropout 0.1 for warmup training on a random 5\% subset of the public dataset for 4 epochs. We apply a learning rate peak of \(2 \times 10^{-5}\) with linear warm-up and cosine decay. Gradients are extracted from each epoch checkpoint and projected to a 1024-dimensional space using Johnson-Lindenstrauss random projections. Data selection is performed by computing cosine similarity between training and validation gradient features aggregated across epochs. The top $N_k^p$ scoring examples are selected for final client $c_k$'s instruction tuning. Additionally, for KnowledgeSG, we utilize AlpacaCare \cite{2024-Zhang-AlpaCare-}, FinGPT \cite{2023-Yang-FinGPT-} and WizardMath \cite{2023-Luo-WizardMath-} as the server-hosted expert model for each domain respectively. To generate the Self-Instruct data, we prompt GPT-3.5 to generate the instruction with the designed prompt in Figure~\ref{fig:instruction}. Specifically, we randomly sample two examples from the client's local data to guide GPT-3.5 generating the in-domain instruction and one example from the client's local data for one-shot in-context learning to guide GPT-3.5 generating responses into the example's format.


\subsection{Effect of Retrieval Number}\label{subsec:retrieval_num}

\begin{figure*}[!tb]
    \centering
    \subfigure[{Different amounts of augmented public data for FedDIT.}]{
        \begin{minipage}[b]{0.35\linewidth}
            \includegraphics[width=\linewidth]{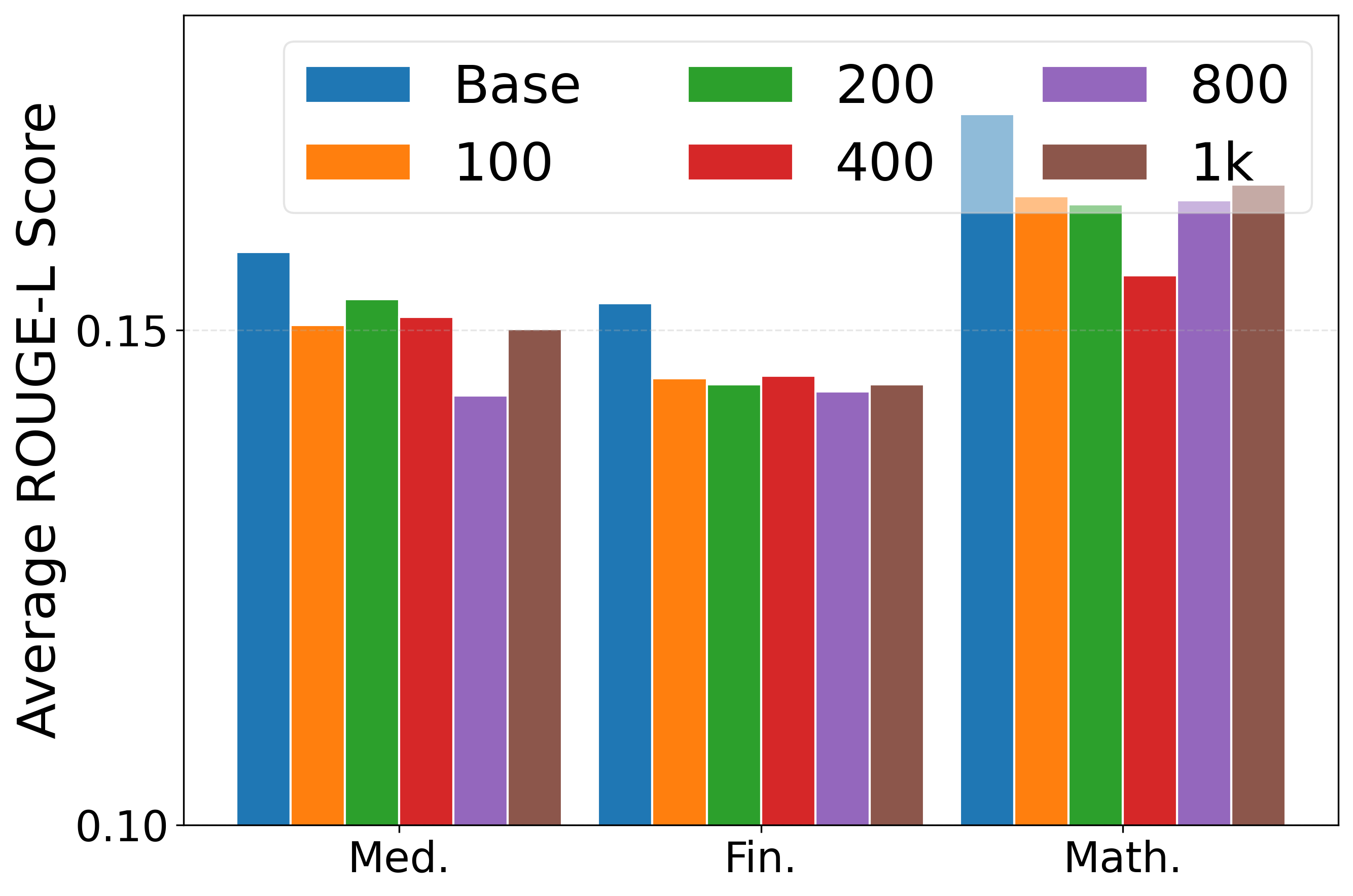}
        \end{minipage}
        \label{subfig:privacy_preserving_different_amount}
    }
    \hspace{8mm}
    \subfigure[{The average ROUGE-L score per round. Dotted lines represent the base-data-only setting.}]{
        \begin{minipage}[b]{0.35\linewidth}  
            \includegraphics[width=\linewidth]{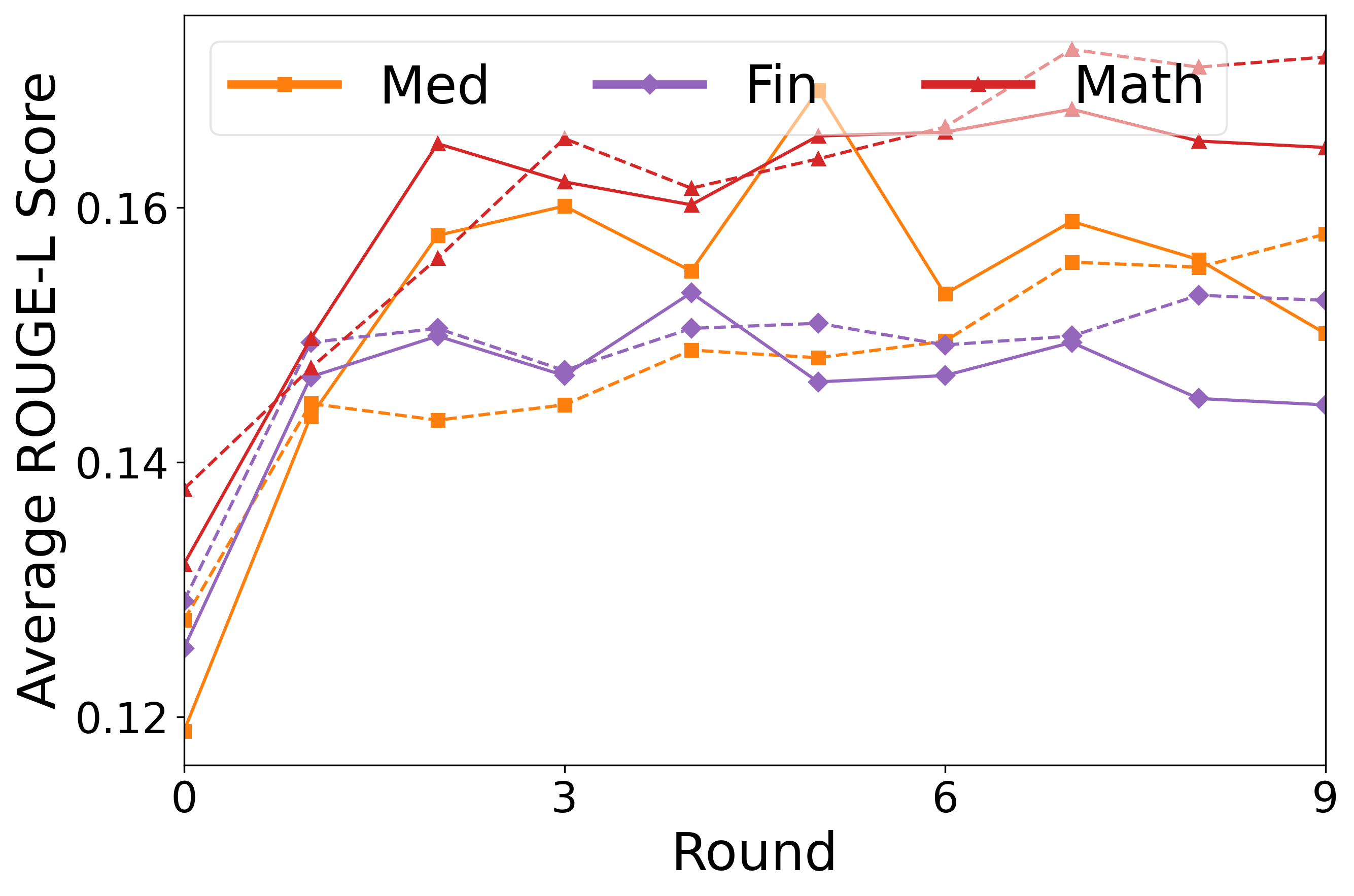}
        \end{minipage}
        \label{subfig:privacy_per_round}
    }
    \vspace{-4mm}  
    \caption{Privacy analysis of memory extraction attacks. Data augmentation can effectively mitigate the privacy leakage risk.}
    \label{fig:privacy}
    \end{figure*}

\paragraph{Experimental Setup.}
To assess the impact of the volume of augmented data, we vary the number of public instructions retrieved per client by FedDCA. Specifically, we test retrieval amounts of [100, 200, 400, 800, 1000] instructions per client, while keeping other experimental parameters consistent with the main setup described in \cref{subsec:experiment_setup}. The experiments are conducted across the three primary domains (medical, financial, and mathematical).

\paragraph{Results.}

\Cref{fig:effect_different_retrieval_numbers} shows FedDCA's performance and domain coverage with varying retrieval amounts. Both generally increase with more retrieved instructions, indicating richer augmentation enhances learning. The improvement rates slow after about 400 instructions per client, likely due to increasing overlap with already covered semantic space or the model reaching its capacity to benefit from further diverse examples under the given training regime. Overall, performance and coverage continue to rise with retrieval volume.

Notably, domain coverage's impact on performance differs by domain. For example, medical domain gains are modest, while financial domain gains are more aligned with coverage increases. This variance may stem from: 1) The base model's initial domain proficiency, which influences its baseline performance and the marginal benefits of additional data. 2) The degree of similarity between the public in-domain data distribution and the test set distribution, where higher relevance yields more proportional performance benefits.

\subsection{Privacy Analysis}\label{subsec:privacy_analysis}
\paragraph{Memory Extraction Attack.} We evaluate the privacy-preserving capability of different ratios of public data against memory extraction attacks \cite{2021-Carlini-Extracting-,2024-Zhang-Effective-}, which utilizes the autoregression nature of LLM\cite{2024-Xu-Magpie-}.

We focus on one client's instruction tuning in FedDIT, using FedDCA for instruction augmentation with 100 to 1k public instructions. We set up 10 clients with full participation for 10 rounds. Specifically, we record the average ROUGE-L score \cite{2004-Lin-ROUGE-TSBO} for client $c_0$ in each round (following the setup in \cite{2024-Zhang-FewFedPIT-}). As we use Llama3-8B as our base model and format the instructions and responses into the Alpaca's format, to utilize the auto-regression nature of LLM to extract the instruction, we prompt the model to generate the instruction using the prompt in Figure~\ref{fig:memory_extraction}, which is exactly the prefix of the Alpaca's template.

For each setting, we repeat memory extraction 100 times and report the average ROUGE-L score. Specifically, denote the generated $\mathcal{N}$ instructions as $\mathcal{I}$ and the client's local instructions $\mathcal{I}^l$. The calculation of the average ROUGE-L score is defined as $\frac{1}{\mathcal{N}} \sum_{i=1}^{\mathcal{N}} \texttt{ROUGE-L} (\mathcal{I}_i,\mathcal{I}^l)$.

\cref{subfig:privacy_preserving_different_amount} shows no significant correlation between the public data ratio and privacy-preserving capability in the same training round. In addition, only using local data has a higher risk of privacy leakage than augmented methods. Additionally, \cref{subfig:privacy_per_round} shows the trends of the average ROUGE-L score per round. Initially, the average ROUGE-L score for augmented settings increases, then decreases or converges, while the base-data-only scores continue to rise, especially in the code domain. This indicates that with more training rounds, base-data-only fine-tuning captures more privacy information, while the privacy leakage risk in augmented fine-tuning decreases or converges.

\paragraph{Domain Inference Attack.} The server may inference the clients' data domain when the domain data retrieval is performed on the server side. However, the proposed algorithm FedDCA is independent of the presence of a public dataset on the server. Even if the server does not have a public dataset, clients can upload their cluster centers to the server, which selects a set of client centers and sends them back to the clients. Each client can then retrieve data from the website based on the received client center by itself, thereby achieving data augmentation while maximizing the cross-client domain coverage. 

In that case, since the server does not know the encoder used by the client, it cannot infer the semantic meaning of the embedding. Thus, for the server, it becomes significantly more challenging to infer the client's domain, let alone apply any privacy protection techniques to the embeddings.

We provide two examples for illustration. Two different encoders are used as client's and server's respectively: \texttt{BAAI/bge-large-en-v1.5} (denoted as $w_1$) and \texttt{google-bert/bert-large-uncased} (denoted as $w_2$). Both encoders output 1024-dimensional features.

\textbf{Example 1:}
Both $w_1$ and $w_2$ take \texttt{``hello world''} as input, and the cosine similarity between their embeddings is \textbf{0.1829}.

\textbf{Example 2:}
Three instructions are used:
\begin{itemize}[itemsep=2pt,topsep=0pt,parsep=0pt]
   \item Instruction 1: \texttt{\small Create an array of length 5 which contains all even numbers between 1 and 10.}
   \item Instruction 2: \texttt{\small Write a replace method for a string class which replaces the given string with a given set of characters.}
   \item Instruction 3: \texttt{\small What is the sentiment of this news? Please choose an answer from \{negative/neutral/positive\}. Teollisuuden Voima Oyj, the Finnish utility known as TVO, said it shortlisted Mitsubishi Heavy's EU-APWR model along with reactors from Areva, Toshiba Corp., GE Hitachi Nuclear Energy, and Korea Hydro \& Nuclear Power Co.}
\end{itemize}

For these instructions, Instruction 1 is passed to $w_1$ and Instructions 2 and 3 are passed to $w_2$, which will result in three embeddings: $e_1$, $e_2$, and $e_3$. The cosine similarity between $e_1$ and $e_2$ is \textbf{0.1464}, while the similarity between $e_1$ and $e_3$ is \textbf{0.1879}. Instructions 1 and 2 are in the same domain, whereas they have a lower cosine similarity.

In conclusion, as demonstrated above, when the clients do not perform domain-specific instruction retrieval on the server side, the server cannot infer the client's domain based on the uploaded embeddings.

\begin{table}[!tb]
    \centering
    \tiny
    \resizebox{0.47\textwidth}{!}{
    \begin{tabular}{@{\hspace{4pt}}l@{\hspace{4pt}}|@{\hspace{4pt}}c@{\hspace{4pt}}c@{\hspace{4pt}}c@{\hspace{4pt}}}
    \toprule
    & MMLU-Med & FPB & GSM8K \\
    \midrule
    Zero-shot & 70.60/- & 55.94/- & 23.27/- \\
    Base Data & 72.40/0.8377 & 66.74/0.9339 & 49.12/0.8709 \\
    Random Sampling & 69.90/0.8497 & 61.05/0.9408 & 47.23/0.8812 \\
    FedDCA & \textbf{73.30}/\textbf{0.9090} & \textbf{67.16}/\textbf{0.9800} & \textbf{50.26}/\textbf{0.9118} \\
    \bottomrule
    \end{tabular}
    }
    \vspace{-2mm}
    \caption{Scalability. Performance (\%) and domain coverage of FedDCA and other baselines. 100 clients with 2 clients per round. Results show the strong scalability of FedDCA.}
    \label{tab:100clients}
    \end{table}

\subsection{Scalability}\label{subsec:scalability}

\begin{figure*}[!tb]
    \centering
    \subfigure[FPB]{
        \begin{minipage}[b]{0.27\textwidth}
            \includegraphics[width=\textwidth]{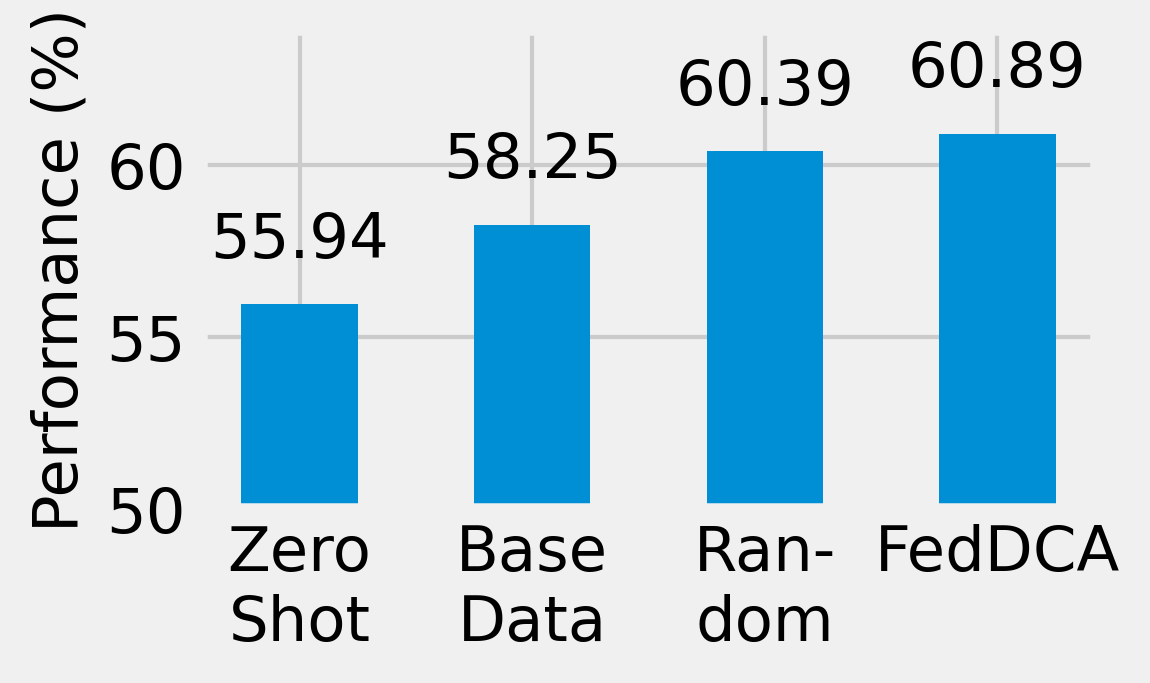}
        \end{minipage}
        \label{subfig:heldout_FPB}
    }\vspace{-1mm}
    \hspace{2mm}
    \subfigure[FiQA]{
        \begin{minipage}[b]{0.27\textwidth}
            \includegraphics[width=\textwidth]{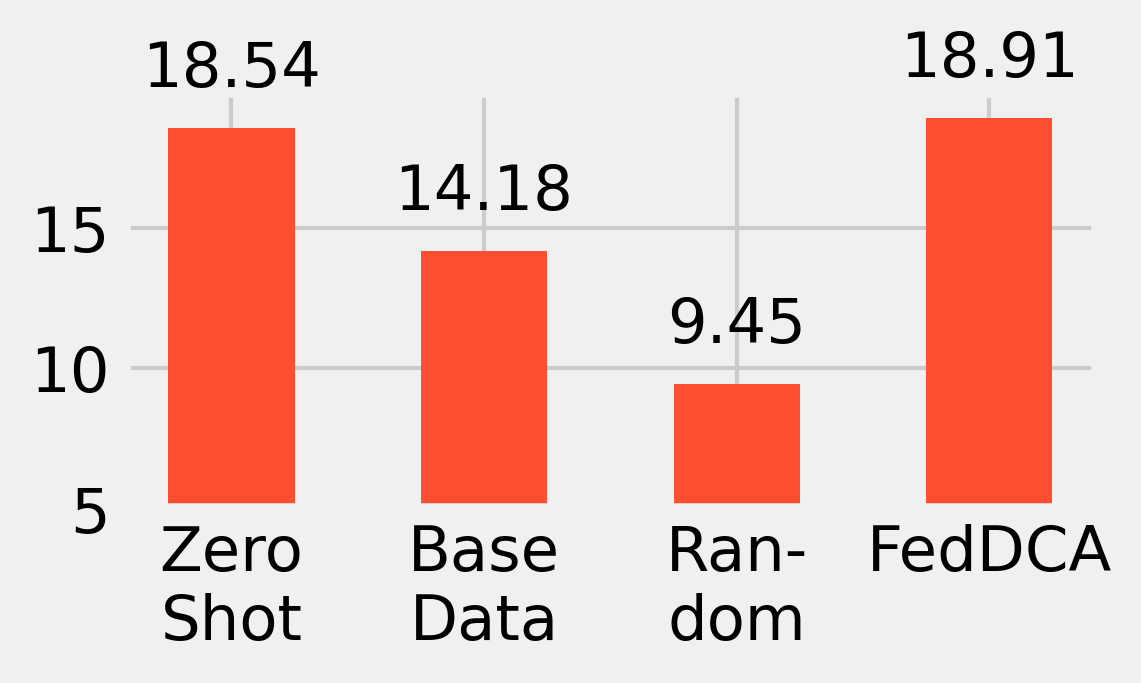}
        \end{minipage}
        \label{subfig:heldout_FiQA}
    }\vspace{-1mm}
    \hspace{2mm}
    \subfigure[TFNS]{
        \begin{minipage}[b]{0.27\textwidth}
            \includegraphics[width=\textwidth]{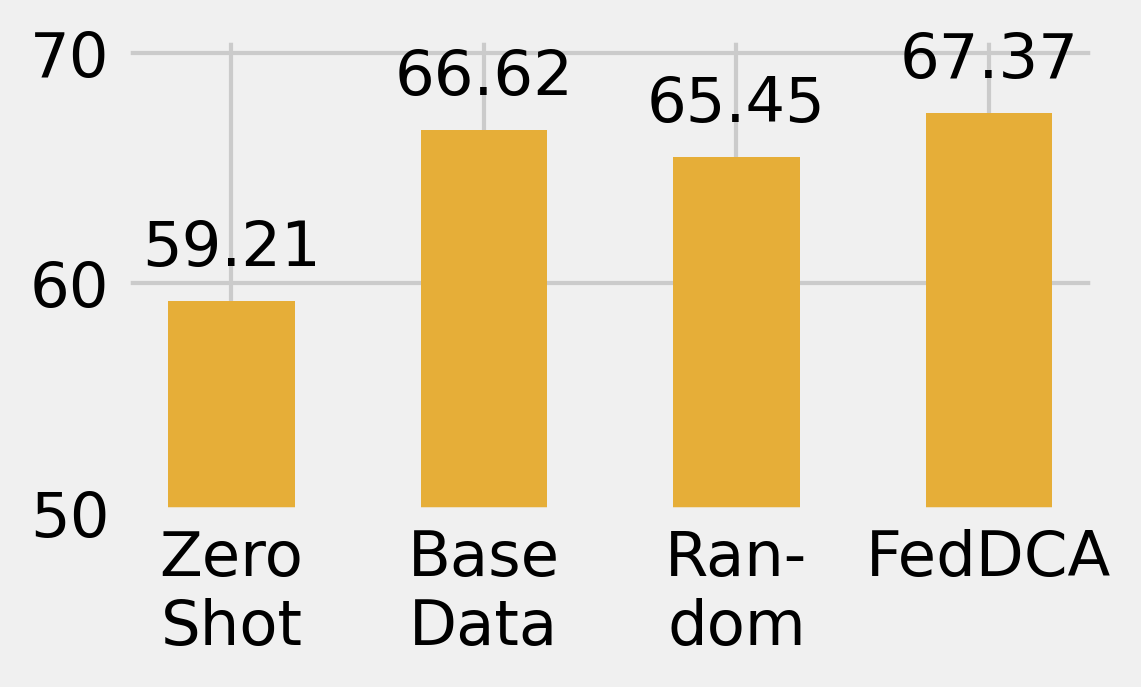}
        \end{minipage}
        \label{subfig:heldout_TFNS}
    }\vspace{-1mm}
    \caption{Held-out setting. We report the performance of FedDCA and other baselines for (a) FPB, (b) FiQA, and (c) TFNS datasets. Results show the robustness of FedDCA in the held-out setting.}
    \end{figure*}
    
\begin{figure}[!tb]
    \centering
    \includegraphics[width=0.35\textwidth]{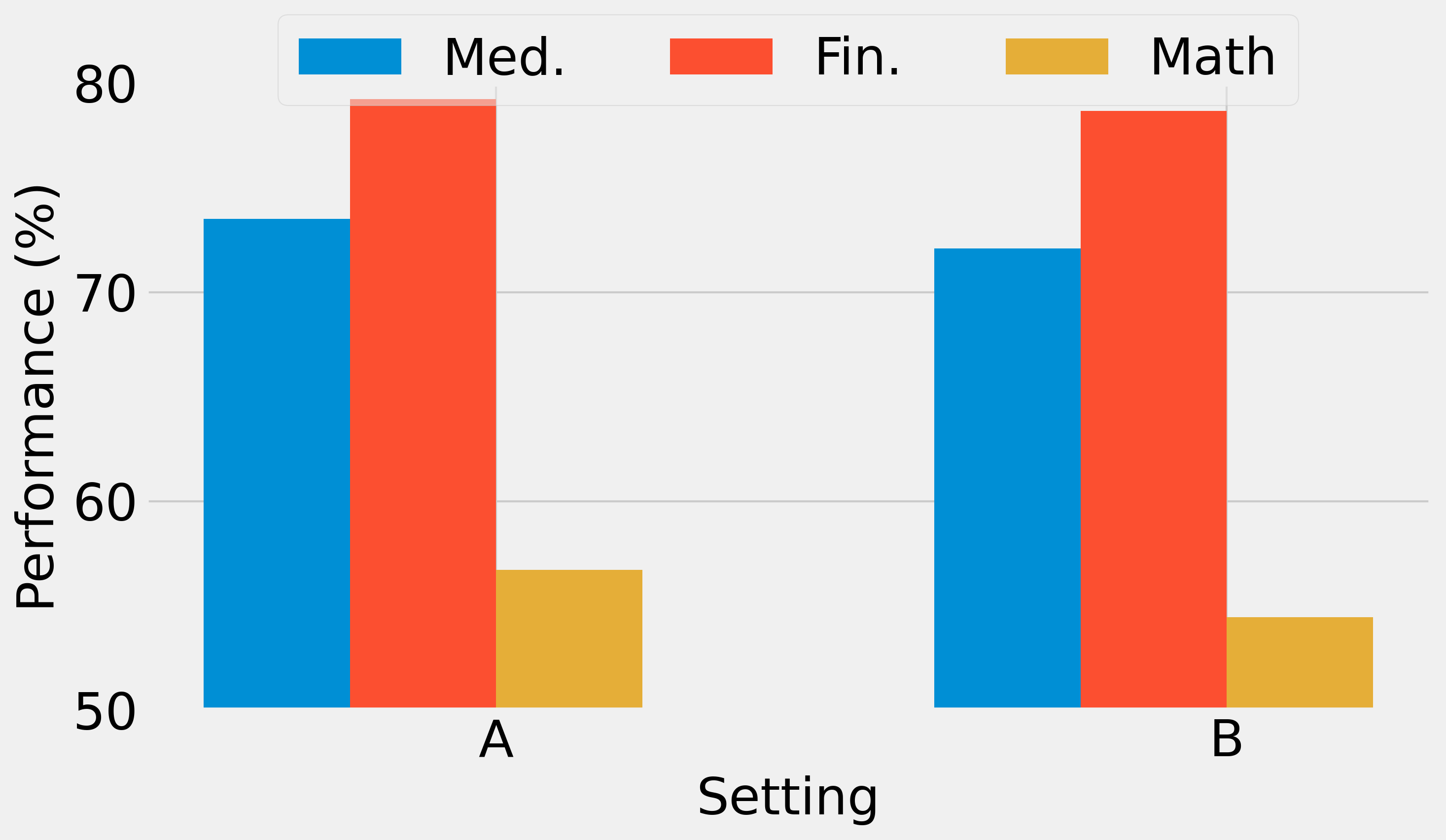}
    \caption{Data selection scenarios. We compare the performance of using the full dataset (Setting A) and using only 10\% of data selected through FedDCA (Setting B), demonstrating FedDCA's efficiency in data utilization.}
    \label{fig:data_selection}
    \end{figure}

\paragraph{Experimental Setup.} To evaluate the scalability of FedDCA, we conduct experiments with 100 clients, where 2 clients are randomly selected for FedDIT in each round. As the number of clients increases, the amount of local data on each client gradually grows, allowing us to assess how FedDCA performs in larger-scale distributed settings.

\paragraph{Results.} \cref{tab:100clients} presents the performance and domain coverage across different domains. Interestingly, we observe that the model trained on only base data even outperforms Random Sampling in domains other than code and narrows the gap with FedDCA. This suggests that with sufficient local data, the benefits of data augmentation become more nuanced. However, FedDCA maintains its advantage by consistently achieving higher domain coverage and better performance across all domains, demonstrating its effectiveness in large-scale federated settings. This superior performance can be attributed to FedDCA's strategic approach to maximizing cross-client domain coverage, which remains crucial even when dealing with a larger overall volume of local data distributed across a greater number of clients.

\subsection{Data Selection Scenarios}\label{subsec:data_selection}

\paragraph{Experimental Setup.} We conduct the experiments on two settings: A) For each domain's FedDIT, we randomly dispatch the in-domain data equally to each client. For example, in the code domain, as the size of the public dataset is 20,022, each client is assigned about 2,000 samples. Then we perform FedDIT on each client's whole data. B) Following A's setting, we first perform FedDCA to determine the client center set, and then we conduct the data selection based on the client center set through dense retrieval to select 200 samples for each client. Finally, we perform FedDIT on the selected data.

\paragraph{Results.} \cref{fig:data_selection} demonstrates that FedDCA can achieve comparable performance to training with the full dataset by using only 10\% data through data selection. This efficiency in data utilization highlights the effectiveness of FedDCA's data selection strategy, which maintains model performance while significantly reducing the required training data.


\subsection{Held-out Setting}\label{subsec:held_out} 
If distributed clients aim to solve tasks based on existing knowledge, the public dataset will inevitably contain knowledge relevant to those domains. This could come from the original corpus (which can be converted into instruction-response pairs by GPT) or from pre-constructed instruction datasets on the website. So the distribution of the public dataset can be categorized as follows: containing \textbf{held-in} or \textbf{held-out} instructions. The held-in indicates that the public dataset contains instructions for the specific task that clients aim to solve, while the held-out indicates that the public dataset does not contain this task's instructions. The paper's default setting is the held-in setting. 

Considering the held-out setting in the financial domain, given that the training set FinGPT and the test sets FPB, FiQA, and TFNS are all related to sentiment analysis tasks. We keep the setting of test sets and replace the FinGPT's instructions in public data with data from a financial QA dataset (\texttt{Sujet-Finance-Instruct-177k}\footnote{\url{https://huggingface.co/datasets/sujet/Sujet-Finance-Instruct-177k}}). The clients' local data are still randomly sampled from FinGPT. This approach yields held-out public data.

\cref{subfig:heldout_FPB,subfig:heldout_FiQA,subfig:heldout_TFNS} shows, FedDCA still achieves performance improvements compared to other baselines for the held-out setting. Additionally, using the Random Sampling data augmentation strategy resulted in performance degradation on the FiQA dataset. This further underscores the necessity of selecting an appropriate data augmentation strategy.

\subsection{Empirical Tightness of FedDCA's Greedy Approximation}\label{subsec:empirical_tightness}

FedDCA's optimization objective is \emph{equivalent to the classic sub-modular facility-location problem}, which guarantees that the simple greedy algorithm attains at least a $1-1/e$ fraction of the optimal value in the worst case. To gauge how tight this bound is in practice, we benchmark the FedDCA algorithm against increasingly stronger (but more expensive) baselines on three domains: Medical, Financial and Mathematical. In particular we position FedDCA between the \textbf{beam search} (with various beam widths) and an \emph{infeasible} \textbf{brute-force solver} that attempts to enumerate all candidate subsets. The beam search, which explores multiple promising branches simultaneously, serves as \textbf{a realistic upper bound} that is strictly better than greedy yet still tractable.

\paragraph{Experimental Setup.}
\begin{enumerate}[itemsep=2pt,topsep=0pt,parsep=0pt]
    \item \textbf{Metric:} Domain coverage $d(D^{d}, \mathcal{P})$ where $\mathcal{P}$ is the selected $N$ client centers, defined as Eq.~\ref{eq:domain_coverage}.
    
    \item \textbf{Baselines:} \textbf{$\mathcal{C}_{\text{all}}$ (default FedDCA)} selects client centers using the union of each client's uploaded cluster center set $\mathcal{C}_{\text{all}}$ as the selection reference to compute marginal gains; \textbf{$\mathcal{P}^*_{C}$} is the unreachable optimum under the same selection reference $\mathcal{C}_{\text{all}}$—marked \emph{timeout}, requiring evaluation of $\binom{100}{10} \approx 1.73 \times 10^{13}$ distinct candidate sets; \textbf{$D^{d}$} uses $D^{d}$ as the selection reference to compute marginal gains, which is often impractical in real-world federated settings (please refer to \cref{subsec:greedy}); \textbf{$\mathcal{P}^*_{D}$} is the optimum under $D^{d}$—also \emph{timeout}; \textbf{Beam search} with widths $256, 512, 1024, 2048$ and, for Medical, $9182$ (does not exceed the baseline $D^{d}$ until width $9182$).
    
    \item \textbf{Approximation ratio:} We report
    \begin{equation}
        \frac{d(D^{d},\mathcal{P}_{\mathcal{C}_{\text{all}}})}{\max_{w} d(D^{d},\mathcal{P}_{\text{beam-}w})} \times 100\%,
    \end{equation}
    where the denominator is the best coverage attained by the beam search. $\mathcal{P}_{\mathcal{C}_{\text{all}}}$ is the selected client center set by FedDCA using $\mathcal{C}_{\text{all}}$ as the selection reference.
\end{enumerate}

\begin{table*}[!tb]
\centering
\small
\resizebox{0.98\textwidth}{!}{
\begin{tabular}{@{\hspace{2pt}}l@{\hspace{2pt}}|@{\hspace{2pt}}c@{\hspace{2pt}}c@{\hspace{2pt}}c@{\hspace{2pt}}c@{\hspace{2pt}}c@{\hspace{2pt}}c@{\hspace{2pt}}c@{\hspace{2pt}}c@{\hspace{2pt}}c@{\hspace{2pt}}c@{\hspace{2pt}}}
\toprule
Domain & $\mathcal{C}_{\text{all}}$ & $\mathcal{P}^*_{C}$ & $D^{d}$ & $\mathcal{P}^*_{D}$ & Beam-256 & Beam-512 & Beam-1024 & Beam-2048 & Beam-9182 & Approx. Ratio \\
\midrule
Med. & 0.6772 & \emph{timeout} & 0.6782 & \emph{timeout} & 0.6776 & 0.6776 & 0.6776 & 0.6778 & 0.6782 & \textbf{99.85\%} \\
Fin. & 0.8379 & \emph{timeout} & 0.8418 & \emph{timeout} & 0.8422 & 0.8427 & 0.8427 & 0.8427 & – & \textbf{99.43\%} \\
Math. & 0.6391 & \emph{timeout} & 0.6699 & \emph{timeout} & 0.7401 & 0.7401 & 0.7404 & 0.7405 & – & \textbf{86.30\%} \\
\bottomrule
\end{tabular}
}
\vspace{-2mm}
\caption{Empirical tightness analysis of FedDCA's greedy approximation. Both theoretical optima ($\mathcal{P}^*_{C}$ and $\mathcal{P}^*_{D}$) are computationally unattainable (\emph{timeout}). The approximation ratio is computed against the best beam search result.}
\label{tab:empirical_tightness}
\end{table*}

\paragraph{Results.}
As shown in \cref{tab:empirical_tightness}, FedDCA using $\mathcal{C}_{\text{all}}$ already achieves \emph{nearly optimal} coverage in Medical and Financial domains, within 99.85\% and 99.43\% of an extensive beam search respectively, and still attains a respectable 86.30\% in the more challenging Mathematical domain.

Note that both theoretical optima ($\mathcal{C}_{\text{all}}$ and $D^d$ as reference) are computationally unattainable (\emph{timeout}), so the beam search constitutes the strongest practical competitor.

In conclusion, these real-world ratios vastly exceed the theoretical lower bound of $1 - 1/e \approx 63\%$, further corroborating that the simple greedy procedure is both \textbf{efficient} and \textbf{effectively near-optimal} for real-world federated deployments.

\section{Robustness to Data Heterogeneity}
\label{sec:robustness_heterogeneity}

The efficacy of retrieval-based augmentation strategies in FedDIT, such as Direct Retrieval and our proposed FedDCA, is inherently linked to the semantic characteristics of clients' local private data, as these inform the selection of public instructions. A critical consideration is how varying degrees of inter-client data heterogeneity impact the ability to construct diverse and comprehensively covering augmented datasets. Intuitively, if clients possess highly similar local data, methods relying on these local signals might struggle to achieve broad cross-client domain coverage and instruction diversity in the augmented sets, which are crucial for a well-generalized global model. 

This section, therefore, investigates the robustness of FedDCA, Direct Retrieval, and Random Sampling to a spectrum of inter-client data heterogeneity. We focus on these methods as their augmentation processes are either directly influenced by local data distributions (FedDCA, Direct Retrieval) or serve as a data-agnostic baseline (Random Sampling); other generative or complex retrieval methods often have different operational dependencies. We control heterogeneity using a Dirichlet distribution ($\text{Dir}(\beta)$) over semantic clusters of clients' private instructions, where smaller $\beta$ values yield higher inter-client heterogeneity (more distinct local data) and larger $\beta$ values lead to lower heterogeneity (more similar local data). We evaluate performance across $\beta \in \{0.01, 0.1, 1.0, 10\}$.

\subsection{Experimental Setup}

\paragraph{Inter-Client Data Heterogeneity Control.}
We focus on the Financial domain, using FinGPT as the source for private instructions.
First, we perform k-means clustering ($\xi=100$ clusters, consistent with \cref{subsubsec:not_matter}) on FinGPT instruction embeddings to create semantic pseudo-labels.
To generate local data for 10 clients, each holding 100 instructions, we employ a Dirichlet distribution $\text{Dir}(\beta)$ over these clusters to define each client's data composition, simulating varying degrees of inter-client data heterogeneity for $\beta \in \{0.01, 0.1, 1.0, 10\}$.

\paragraph{Augmentation and Training.}
Following the general setup in \cref{subsec:experiment_setup}, Random Sampling, Direct Retrieval (introduced in \cref{sec:preliminaries}), and \mbox{FedDCA} are used to obtain 1k augmented instructions per client before commencing FedDIT.

\paragraph{Evaluation Metrics.}
Model performance is evaluated using FPB, FiQA, and TFNS accuracy.
Augmented data characteristics are assessed using three key metrics:

1) \textbf{Domain Coverage ($\uparrow$)}: As defined in Eq.~\ref{eq:domain_coverage}, this measures how well the aggregated client data (local + augmented) $D^c$ covers the target financial domain $D^d$.

2) \textbf{ICACS (Inter-Client Augmented Centroids Similarity $\downarrow$)}: For each client $k$, its augmented instruction embeddings are clustered (k-means, $\xi=10$) into centers $\mathcal{C}_k$. ICACS is the average pairwise cosine similarity between all such cluster centers across different clients. Lower values indicate more distinct augmented sets at a granular level.

3) \textbf{RUAI (Ratio of Unique Augmented Instructions $\uparrow$)}: The ratio of unique instructions within the total cross-client augmented data pool $D^c$. Higher values signify less redundancy.

\vspace{-1mm}
\subsection{Results and Analysis}
\vspace{-1mm}

\begin{table*}[htbp]
    \centering
  \resizebox{\textwidth}{!}{%
  \begin{tabular}{@{}lcrrrccc@{}}
    \toprule
    Dirichlet $\beta$ & Augmentation & \multicolumn{3}{c}{Model Performance (\%)} & \multicolumn{3}{c}{Augmented Data Characteristics} \\
    \cmidrule(lr){3-5} \cmidrule(lr){6-8}
     & Strategy   & FPB    & FiQA   & TFNS   & Domain Coverage ($\uparrow$) & ICACS ($\downarrow$) & RUAI ($\uparrow$) \\
    \midrule
    \multirow{3}{*}{0.01 } 
        & Random Sampling & 64.20  & 13.10  & 65.55  & 0.920 & 0.472 & 0.867 \\
        & Direct Retrieval& 67.20  & 21.50  & 69.00  & 0.938 & 0.434 & 0.909 \\
        & FedDCA          & \textbf{67.90} & \textbf{36.30} & \textbf{74.40} & \textbf{0.985} & \textbf{0.405} & \textbf{0.943} \\
    \midrule
    \multirow{3}{*}{0.1} 
        & Random Sampling & 64.19  & 13.09  & 65.53  & 0.920 & 0.472 & 0.867 \\
        & Direct Retrieval& 66.31  & 19.11  & 67.62  & 0.929 & 0.441 & 0.882 \\
        & FedDCA          & \textbf{67.24} & \textbf{35.27} & \textbf{73.32} & \textbf{0.982} & \textbf{0.411} & \textbf{0.938} \\
    \midrule
    \multirow{3}{*}{1.0}
        & Random Sampling & 64.19  & 13.09  & 65.53  & 0.919 & 0.472 & 0.867 \\
        & Direct Retrieval& 65.50  & 17.80  & 66.79  & 0.924 & 0.448 & 0.879 \\
        & FedDCA          & \textbf{66.94} & \textbf{34.82} & \textbf{72.66} & \textbf{0.977} & \textbf{0.414} & \textbf{0.927} \\
    \midrule
    \multirow{3}{*}{10}  
        & Random Sampling & 64.20  & 13.05  & 65.45  & 0.918 & 0.472 & 0.867 \\
        & Direct Retrieval& 64.37  & 15.20  & 66.10  & 0.922 & 0.452 & 0.871 \\
        & FedDCA          & \textbf{66.73} & \textbf{34.51} & \textbf{72.23} & \textbf{0.972} & \textbf{0.423} & \textbf{0.921} \\
    \bottomrule
  \end{tabular}
  }
  \caption{Performance (\%), domain coverage, ICACS and RUAI across different levels of client data heterogeneity (controlled by Dirichlet parameter $\beta$) in the Financial Domain. Best model performance, domain coverage, ICACS and RUAI per $\beta$ setting are in \textbf{bold}.}
  \label{tab:robustness_heterogeneity}
\end{table*}


\Cref{tab:robustness_heterogeneity} systematically quantifies the impact of inter-client data heterogeneity on augmentation strategies. For small $\beta$ (i.e., high heterogeneity), the cross-client data exhibits substantial diversity, enabling Direct Retrieval to access a broader range of semantic regions in the public pool. This diversity translates into clear gains over Random Sampling in both performance (e.g., +6\% FiQA at $\beta=0.01$) and coverage metrics. However, as $\beta$ increases and client data distributions become more homogeneous, the advantage of Direct Retrieval diminishes, with its performance and diversity metrics converging toward those of Random Sampling. This trend underscores that the effectiveness of Direct Retrieval is fundamentally driven by the diversity present in cross-client data.

Conversely, as inter-client heterogeneity decreases (i.e., local data distribution becomes more similar), Direct Retrieval's effectiveness continuously degrades. Its model performance and domain coverage drop, showing a more sensitive effectiveness to the heterogeneity. Results indicate that with more homogeneous local data distributions, Direct Retrieval tends to fetch more similar public instructions due to the similarity between each client's cluster centers. 

Random Sampling exhibits stable, albeit consistently sub-optimal, performance and diversity metrics across all heterogeneity levels. This stability stems from its sampling strategy (oblivious to local data distributions) of the public dataset, which results in similar augmented data.

Additionally, \mbox{FedDCA} showcases strong robustness. It consistently achieves the best model performance across all tested $\beta$ values. While its diversity metrics (domain coverage, ICACS, RUAI) show minor sensitivity to the cross-client data distribution (optimally diverse at highest heterogeneity). This resilience stems from its coverage-oriented client center selection (\cref{alg:greedy_center_selection}), which effectively prevents redundant augmentation by selecting public instructions that add novel semantic aspects to the collective, even with similar cross-client data distributions. In conclusion, these findings underscore \mbox{FedDCA}'s robustness and adaptability irrespective of the specific degree of inter-client heterogeneity, makes it a more reliable and effective solution for real-world FedDIT deployments.

\section{Augmentation Strategy Visualization}\label{subsec:augmentation_strategy_visualization}
\begin{figure*}[htbp]
\centering
\subfigure[Medical]{
    \begin{minipage}[b]{0.6\textwidth}
        \includegraphics[width=1\textwidth]{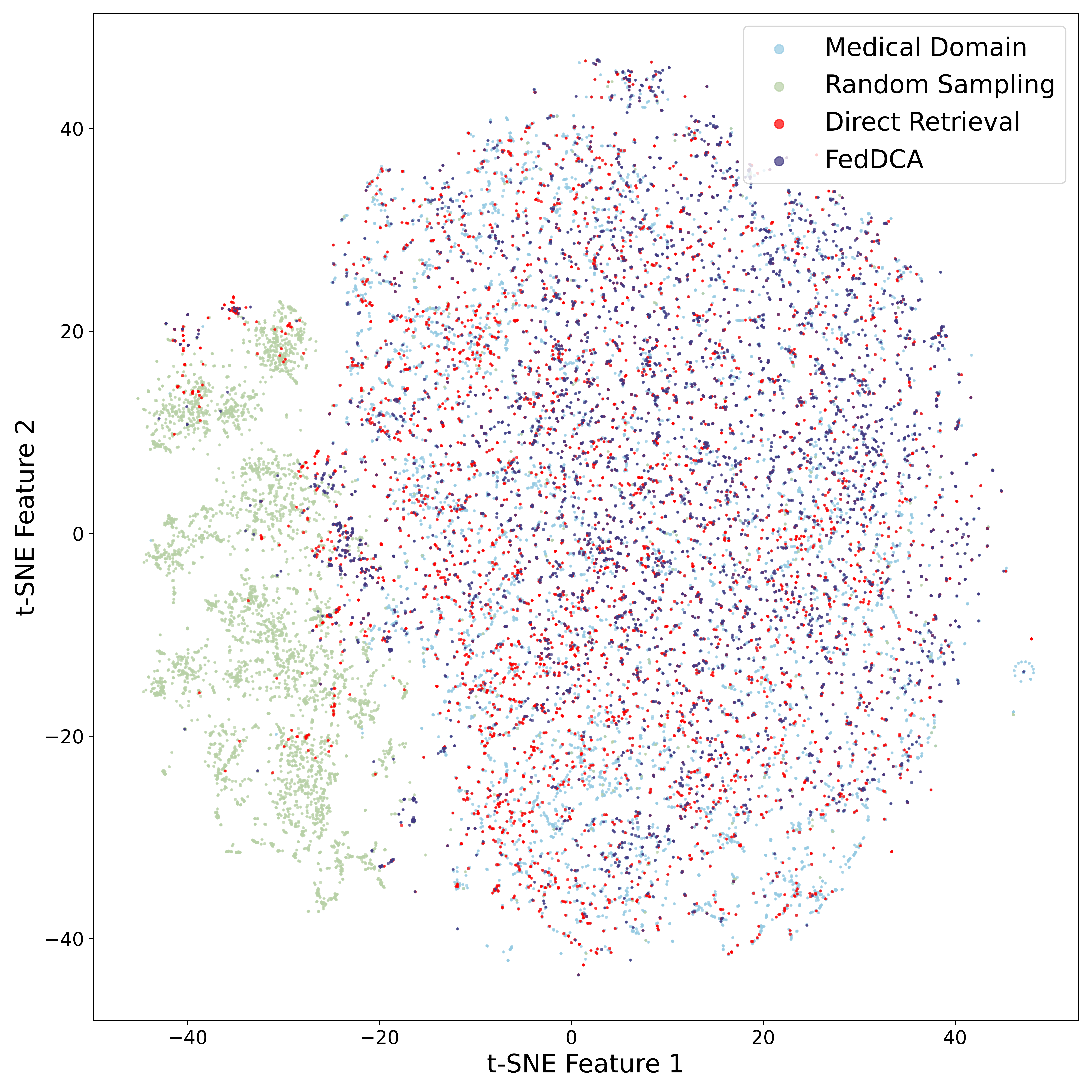}
    \end{minipage}
}
\vspace{2mm}
\subfigure[Financial]{
    \begin{minipage}[b]{0.45\textwidth}
        \includegraphics[width=1\textwidth]{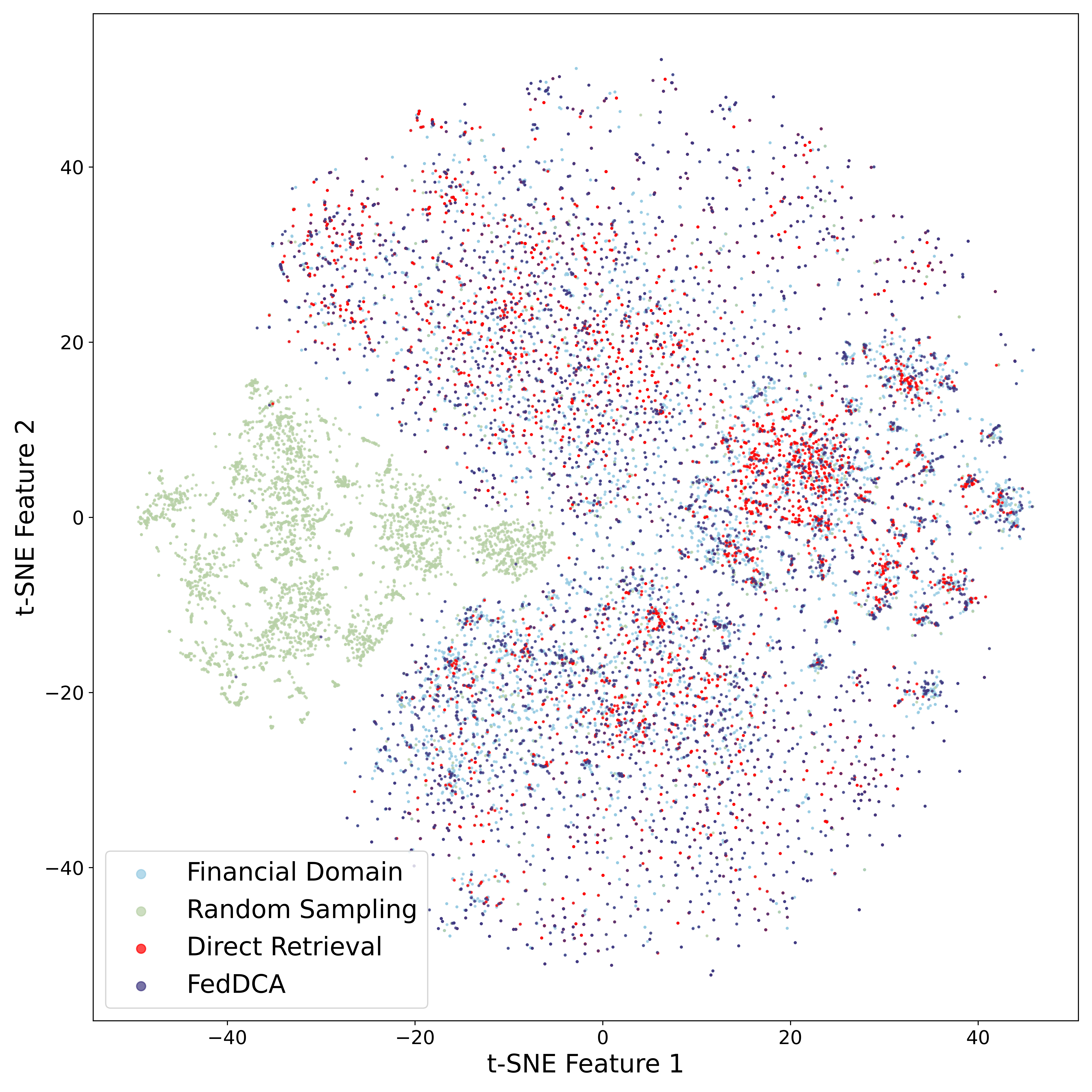}
    \end{minipage}
}
\vspace{2mm}
\subfigure[Mathematical]{
    \begin{minipage}[b]{0.45\textwidth}
        \includegraphics[width=1\textwidth]{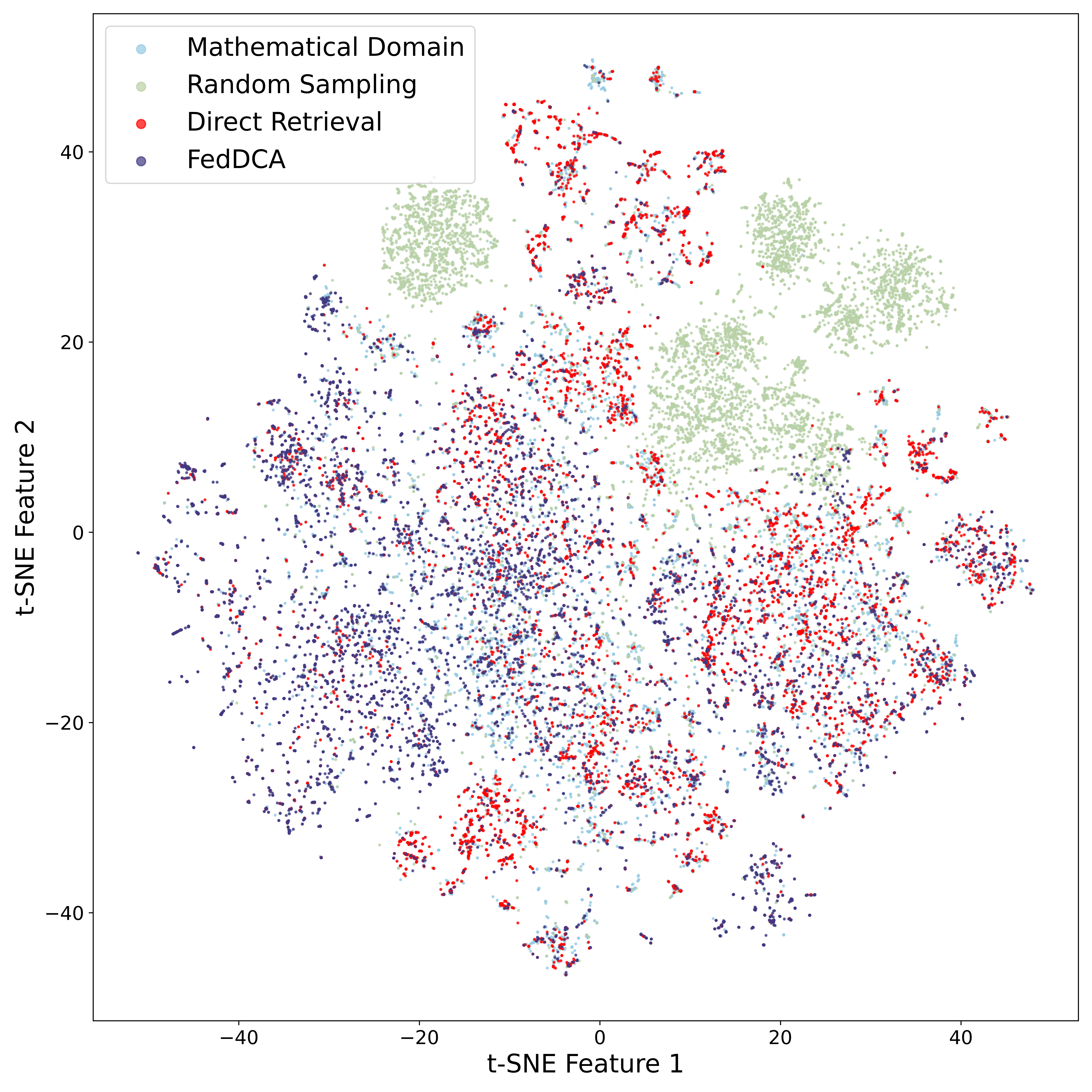}
    \end{minipage}
}
\caption{Visualization of cross-client data distribution in different domains, performing t-SNE dimensionality reduction on retrieved instructions through various augmentation strategies. We randomly sample 10k in-domain samples as background while randomly sampling 5k samples from the cross-client augmented dataset for different instruction augmentation methods for comparison.}
\label{fig:augmentation_strategy_visualization}
\end{figure*}

To more intuitively compare the domain coverage of different instruction augmentation methods, we randomly sample 5k cross-client instructions obtained through these methods and 10k in-domain instructions as the background, representing the distribution of specific domains in the public dataset. We then visualized the results using t-SNE \cite{2008-Maaten-Visualizing-JMLR}, as shown in Figure~\ref{fig:augmentation_strategy_visualization}. The plot shows that FedDCA encompasses most of the in-domain data, which is consistent with FedDCA's domain coverage of each domain shown in Table~\ref{tab:performance}. Also, we can observe that the random sampling strategy selects a lot of out-of-domain data while does not have good coverage in specific domains.

\end{document}